%% file: sigir22-sigconf.tex
\documentclass[sigconf,natbib=true]{acmart}

\usepackage{multirow}
\usepackage{booktabs}
\usepackage{natbib}
\usepackage[ruled,vlined]{algorithm2e}
\usepackage{algpseudocode}
\usepackage[utf8]{inputenc}
\usepackage{amssymb,amsmath,caption}

\usepackage[hang,flushmargin]{footmisc}
\usepackage{lipsum}
\usepackage{listings}
\usepackage{extarrows}
\usepackage{xspace}
\usepackage{graphicx}
\usepackage{subcaption}
\usepackage{color}

\usepackage{makecell}
\usepackage{ifthen}
\usepackage{xifthen}
\usepackage{wrapfig}
\usepackage{amsthm}
\usepackage{CJKutf8}

\definecolor{PineGreen}{RGB}{34,139,34}
\newtheorem{definition}{Definition}
\newcommand{\solution}[0]{ICLEA\xspace}

\AtBeginDocument{%
  \providecommand\BibTeX{{%
    \normalfont B\kern-0.5em{\scshape i\kern-0.25em b}\kern-0.8em\TeX}}}

\setcopyright{acmcopyright}

\copyrightyear{2022} 
\acmYear{2022} 
\setcopyright{rightsretained} 

\acmConference[CIKM '22]{Proceedings of the 31st ACM International Conference on Information and Knowledge Management}{October 17--21, 2022}{Atlanta, GA, USA}
\acmBooktitle{Proceedings of the 31st ACM Int'l Conference on Information and Knowledge Management (CIKM '22), Oct. 17--21, 2022, Atlanta, GA, USA}
\acmDOI{10.1145/3511808.3557364}
\acmISBN{978-1-4503-9236-5/22/10}

\usepackage{etoolbox}
\makeatletter
\patchcmd{\maketitle}{\@copyrightpermission}{
   \begin{minipage}{0.3\columnwidth}
     \href{https://creativecommons.org/licenses/by/4.0/}{\includegraphics[width=0.90\textwidth]{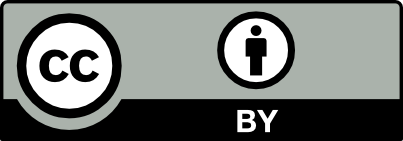}}
   \end{minipage}\hfill
   \begin{minipage}{0.7\columnwidth}
     \href{https://creativecommons.org/licenses/by/4.0/}{This work is licensed under a Creative Commons Attribution International 4.0 License.}
   \end{minipage}
  
   \vspace{5pt}
}{}{}

\makeatother

\begin{document}

\title{Interactive Contrastive Learning for Self-Supervised \\Entity Alignment}
%
\author{Kaisheng Zeng}
\email{zks19@mails.tsinghua.edu.cn}
\orcid{0000-0002-8104-9652}
\affiliation{%
  \institution{Tsinghua University}
  \streetaddress{30 Shuangqing Rd}
  \state{Beijing}
  \country{China}}

\author{Zhenhao Dong}
\email{dzh0129@outlook.com}
\affiliation{%
  \institution{Beijing University of Chemical Technology}
  \state{Beijing}
  \country{China}}

\author{Lei Hou}
\authornote{Corresponding author.}
\email{houlei@tsinghua.edu.cn}
\affiliation{%
  \institution{Department of Computer Science and Technology, BNRist, Tsinghua}
  \city{Beijing}
  \country{China}
}

\author{Yixin Cao}
\affiliation{%
  \institution{Singapore Management University}
  \country{Singapore}}
  
\author{Minghao Hu}
\affiliation{%
  \institution{Information Research Center of Military Science}
  \state{Beijing}
  \country{China}}

\author{Jifan Yu}
\author{Xin Lv}
\affiliation{%
  \institution{Tsinghua Univerisity}
  \city{Beijing}
  \country{China}
}

\author{Lei Cao}
\affiliation{%
  \institution{Tsinghua University}
  \city{Beijing}
  \country{China}
}

\author{Xin Wang}
\affiliation{%
  \institution{Tsinghua University}
  \city{Beijing}
  \country{China}
}

\author{Haozhuang Liu}
\affiliation{%
  \institution{Tsinghua University}
  \city{Beijing}
  \country{China}
}


\author{Yi Huang}
\author{Junlan Feng}
\affiliation{%
  \institution{China Mobile Research Institute}
  \city{Beijing}
  \country{China}
}

\author{Jing Wan}
\affiliation{%
  \institution{Beijing University of Chemical Technology}
  \state{Beijing}
  \country{China}}
 

\author{Juanzi Li}
\author{Ling Feng}
\affiliation{%
  \institution{CST, BNRist, Tsinghua University}
  \city{Beijing}
  \country{China}
}

\renewcommand{\shortauthors}{Kaisheng Zeng et al.}
\renewcommand{\shorttitle}{Interactive Contrastive Learning for Self-Supervised Entity Alignment}

\begin{abstract}

Self-supervised entity alignment (EA) aims to link equivalent entities across different knowledge graphs (KGs) without the use of pre-aligned entity pairs. The current state-of-the-art (SOTA) self-supervised EA approach draws inspiration from contrastive learning, originally designed in computer vision based on instance discrimination and contrastive loss, and suffers from two shortcomings. Firstly, it puts unidirectional emphasis on pushing sampled negative entities far away rather than pulling positively aligned pairs close, as is done in the well-established supervised EA. Secondly, it advocates the minimum information requirement for self-supervised EA, while we argue that self-described KG's side information (e.g., entity name, relation name, entity description) shall preferably be explored to the maximum extent for the self-supervised EA task. In this work, we propose an interactive contrastive learning model for self-supervised EA. It conducts bidirectional contrastive learning via building pseudo-aligned entity pairs as pivots to achieve direct cross-KG information interaction. It further exploits the integration of entity textual and structural information and elaborately designs encoders for better utilization in the self-supervised setting. Experimental results show that our approach outperforms the previous best self-supervised method by a large margin (over 9\% Hits@1 absolute improvement on average) and performs on par with previous SOTA supervised counterparts, demonstrating the effectiveness of the interactive contrastive learning for self-supervised EA. The code and data are available at \url{https://github.com/THU-KEG/ICLEA}.

\end{abstract}





\maketitle

\input{paragraphs/1.introduction0}

\input{paragraphs/2.related_work}

\input{paragraphs/2.define}
\input{paragraphs/3.method}
\input{paragraphs/4.experiments}

\input{paragraphs/5.conclusion}
\input{paragraphs/7.acknowledgement}

\bibliographystyle{ACM-Reference-Format}
\bibliography{sigir22-sigconf}

\appendix








\end{document}

%% file: paragraphs/1.introduction0.tex
\section{Introduction}

Knowledge Graphs (KGs) (e.g., DBpedia~\cite{z3_lehmann2015dbpedia}, YAGO~\cite{mahdisoltani2014yago3}, and Wikidata~\cite{vrandevcic2014wikidata}) provide structural knowledge about the entities and relations in the real world. These separately constructed KGs contain heterogeneous but complementary knowledge~\cite{JAPE}. Entity Alignment (EA) integrates the complementary knowledge in these KGs via identifying equivalent entities~\cite{sun13benchmarking} and thus benefits various knowledge-driven applications such as information extraction~\cite{z23_han2018neural}, question answering~\cite{z11_cui2019kbqa}, and recommendation system~\cite{cao2019unifying,fu2020fairness_sigir,chen2020jointly_sigir}.

The mainstream solutions in the literature are based on deep representation learning,
which embeds entities into a latent space, and then calculates the distance between embeddings
as the evidence for EA. These approaches~\cite{MTransE,GCN-Align,liu2020exploring,tang2019bert-int} utilize seed alignments to guide the representation learning of entity and relation and propagate those seed alignments to the entire KG. Despite the excellent performance, they rely on some 
pre-aligned entity pairs provided by humans during the training process.
As the acquisition of seed alignments is usually time-consuming, labor-intensive, and error-prone, it is hard to apply them to real EA-related scenarios~\cite{li2008rimom}.

Hence, self-supervised EA under the \emph{unsupervised} setting 
(i.e., matching entities without any pre-aligned label) starts to attract research attention. 
One pioneer work is EVA~\cite{liu2021visual_eva}, which used visual semantic representations of entities 
to align entities in heterogeneous KGs in a fully unsupervised setting.
Very recently, SelfKG~\cite{liu2021self} drew inspiration from contrastive learning approaches, 
which originally targeted at computer vision tasks by leveraging instance discrimination and contrastive loss
(e.g., MoCo~\cite{he2020momentum}, SimCLR~\cite{chen2020simple}), 
and provided a contrastive learning framework for self-supervised entity alignment.
Although SelfKG achieved a promising performance, it suffers from two shortcomings. 

First, it struggles with the dilemma of precise entity alignment situation, because it puts unidirectional emphasis on pushing sampled negative entities far away for the uniformity of the whole representation space rather than pulling positively aligned pairs close to enhance the alignment. Its self-negative sampling strategy can only sample entities from the same source KG to avoid conflict, which blocks the direct cross-KG information interaction. EA is essentially building inter-KG links, and thus the cross-KG information interaction is critical. Existing methods achieve such interaction via explicit supervision (seed alignments~\cite{tang2019bert-int}) or implicit signals (images as visual pivots~\cite{liu2021visual_eva}). We argue that pushing sampled negative entities away
and pulling possibly aligned entities close are equally important. 
The latter is actually the core in  
well-established supervised EA approaches. 
 
\begin{figure}[t]
\centering
\includegraphics[width=1.0\linewidth]{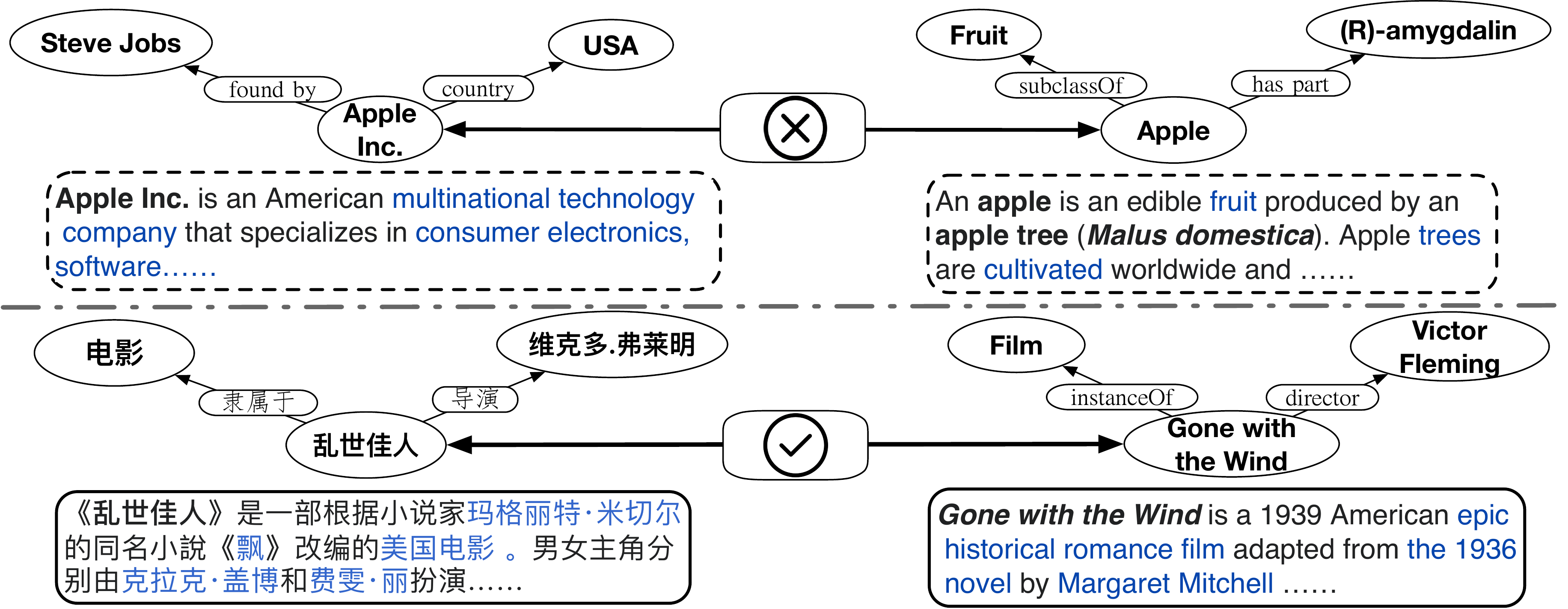}
\caption{Examples of KG's side information (e.g., entity description) enhancing EA when entities' names are deceptive.}
\label{fig:example_desc}
\end{figure}

Second, it follows the minimum information requirement for self-supervised EA and only focuses on entity names, which ignores rich entity semantics in KGs. As shown in Figure~\ref{fig:example_desc}, the entity relationship and description provide rich information, while sometimes relying only on entity names can introduce ambiguity problems. Entity name ambiguity problem~\cite{han2009named} brings severe challenges to self-supervised EA; as shown in Figure~\ref{fig:example_desc}, entity names can be deceptive to EA while KG's side information containing extensive entity semantics may bring great benefit. So we argue that self-described KG's side information (e.g., entity name, relation name, entity description) shall 
preferably be explored to the maximum extent for the self-supervised EA task.

To validate the above two hypotheses, we propose a model named Interactive Contrastive Learning for self-supervised Entity Alignment (\textbf{\solution}). We design a novel interactive contrastive learning mechanism in a self-supervised EA framework by constructing pseudo-aligned entity pairs as virtual pivots to establish a direct information interaction channel for the two KGs. Inspired by previous supervised approaches, we introduce more important side information into self-supervised EA. Specifically, we separately encode entity name labels and descriptions by different pre-trained models, then organically combine them to provide powerful initial embeddings for entities. We also propose a relation-aware neighborhood aggregator to better leverage the structural and semantic information brought by the KGs' relations. 
 
The main contributions of our work are threefold:    
\begin{itemize}
  \item We propose an interactive contrastive learning to achieve the direct cross-KG information interaction via building pseudo-aligned entity pairs in a self-supervised EA.
  \item We propose to better utilize and integrate the structural and semantic entity information for a self-supervised EA framework with the help of pre-trained language models.
  \item Experimental results show that \solution outperforms the best self-supervised baseline by a large margin (over 9\% Hits@1 absolute improvement on average), and performs on par with previous SOTA supervised methods while maintaining more stable model training. Our work significantly narrows the gap between supervised and self-supervised EA approaches.
\end{itemize}

%% file: paragraphs/2.related_work.tex
\section{Related Work}
\label{sec:related_work}
We discuss two lines of work that are relevant to ours according to their usage of pre-aligned entity pairs during the model's training process: (semi-)supervised and self/un-supervised approaches. 
\subsection{(Semi-)Supervised EA Approaches}
In (semi-)supervised approaches, pre-aligned entity pairs are provided during the training process. Early EA methods focused on Bayesian-based probability estimation~\cite{tang2006using} and hand-crafted relevant similarity factors~\cite{li2008rimom}. 
Before the wave of representation learning arrived, most EA approaches either relied on extra resources~\cite{wang2013boosting} or human efforts~\cite{mahdisoltani2014yago3} to discover identical entity pairs.

The deep embedding-based EA approaches seek to encode a KG into a low-dimensional latent space. Earlier approaches focus on how to handle this task using only the structure of KG~\cite{sun13benchmarking,zeng2021comprehensive}. MTransE~\cite{MTransE} uses TransE to learn the vector space of a single KG and then learns a linear transformation to align two vector spaces. 
IPTransE~\cite{iptranse} models relational paths by inferring the equivalence between direct relationships and multi-hop paths. DAT~\cite{zeng2020degree} offers a framework for EA with emphasis on long-tail entities. GNNs~\cite{bruna2013spectral,defferrard2016convolutional,velivckovic2017graph} are well suited for modeling graph structure, and have been widely used for modeling the multifaceted information in KGs to assist embedding-based EA recently~\cite{wang2018cross,MuGNN,li2019semi,zhu2020relation,zhu2020collective,mao2020mraea,sun2020knowledge,liu2020exploring,zeng2021encoding,yu2021generalized}. The emergence of pre-trained language models (PLMs)~\cite{devlin2018bert} has recently brought natural language processing to a new era. Some PLMs can bring high-quality multilingual initial embeddings, which is crucial for the EA task. HMAN~\cite{wu2019jointly} uses multilingual BERT to initialize the embedding of entities by its side information which bridges multilingual gaps. BERT-INT~\cite{tang2019bert-int} proposes an EA strategy based on directly fine-tuned multilingual BERT. Besides the high performance achieved by these (semi-) supervised models, researchers begin to consider the efficiency and practicality with new methods such as adversarial learning, active learning, etc~\cite{pei2020rea,zeng2021reinforced,mao2021negative,liu2021activeea}.
Despite the success of these approaches, the imperative need for labeled data still bridges the gap between these models and real-world applications. The acquisition of pre-aligned seeds is naturally time-consuming and labor-intensive~\cite{liu2020self}. To address this problem, self-supervised or unsupervised EA has gained increasing attention from the academic community.

\subsection{Self-Supervised or Unsupervised EA Approaches}
In self-supervised or unsupervised\footnote{Self-supervised learning is a form of unsupervised learning. Their distinction is informal in the existing literature. In this work, we do not distinguish between them in a strict sense.}~\cite{liu2020self,jing2020self} approaches, none of the pre-aligned entity pairs is provided during the training process. MultiKE~\cite{zhang2019multi} designs some cross-KG inference methods to enhance the alignment between two KGs. EVA~\cite{liu2021visual_eva} uses images as pivots for aligning entities in different KGs in an unsupervised setting.  

Self-supervised learning methods are proposed to learn data co-occurrence relationships from unlabeled data without using any human-annotated labels~\cite{jing2020self}. Self-supervised image feature learning in the computer vision (CV) community has obtained great success. The margin between self-supervised methods' performance and supervised methods' performance on some downstream tasks becomes very small~\cite{he2020momentum,chen2020simple}. Strikingly, the success of self-supervised training in computer vision community provides new directions for unsupervised EA models.
As there are few attempts that launch self-supervised training in this task, such as SelfKG~\cite{liu2021self}, we recognize that it is urgent and purposeful to set up a paradigm so that the experience of existing EA methods can be adapted to this new direction. Specifically, SelfKG designs a unidirectional contrastive learning strategy between two KGs for self-supervised EA and uses self-negative sampling to sample entities from the same source KG to avoid conflict, which blocks the direct cross-KG information interaction. Therefore, we propose ICLEA, which jointly considers the structural and semantic information in the self-training process and constructs pseudo-aligned entity pairs as virtual pivots to guide the direct bidirectional cross-KG information interaction, and calls for more efforts in self-supervised EA.

%% file: paragraphs/2.define.tex
\section{Problem Definition}
\label{sec:problem_defi}

\begin{definition}[Knowledge Graph]
A knowledge graph is represented as $\mathcal{G}$=$(\mathcal{E}, \mathcal{R}, \mathcal{T}, \mathcal{S})$, where each $e \in \mathcal{E} $, $r \in \mathcal{R}$, $t = (e_i, r, e_j ) \in \mathcal{T}$ ( $e_i, e_j \in \mathcal{E} $ ) represent an entity, a relation and a fact triple respectively, and $s(e)=\left \{ n_e,d_e,a_e \right \} \in \mathcal{S}$ denotes the side information of entity $e$, i.e., entity name label, textual description and attribute value. We denote the set of one-hop neighbors of entity $e$ as $\mathcal{N}_{e}$ of size $\left | \mathcal{N}_{e} \right |$, namely, the entities that are directly connected to $e$ in knowledge graph $\mathcal{G}$ via fact triples, where $\mathcal{N}_{e} = \{e^{\prime} ~|~ (e, r, e^{\prime}) \in \mathcal{T} \wedge (e, e^{\prime} \in \mathcal{E})\} \cup \{e^{\prime} ~|~ (e^{\prime}, r, e) \in \mathcal{T} \wedge (e, e^{\prime} \in \mathcal{E})\}$. 
\end{definition}

Knowledge graphs are often separately constructed for various goals, and thus contain heterogeneous but complementary knowledge. Entity alignment is to identify entities from different knowledge graphs (different languages or sources) that describe the same real-world object, and can be formally defined as follows.
\begin{definition}[Entity Alignment]
Given two different KGs $\mathcal{G}_1$ and $\mathcal{G}_2$, EA is to learn a ranking function $\mathit{f}: \mathcal{E}_1 \times \mathcal{E}_2 \rightarrow \mathbb{R}$ to calculate the similarity score between two entities, based on which we rank the correctly aligned entity $e_2$ as high as possible among all entities of $\mathcal{E}_2$ with a queried entity $e_1 \in \mathcal{E}_1$. Pre-aligned entity pairs or named seed alignments $\mathcal{I} = \left \{ (e_1,e_2) \mid e_1 \in \mathcal{E}_1, e_2 \in \mathcal{E}_2,  e_{1} \leftrightarrow e_{2} \right \}$ are provided, where $\leftrightarrow$ means that the entity $e_{1}$ from $\mathcal{G}_{1}$ and the entity $e_{2}$ from $\mathcal{G}_{2}$ are equivalent.
\end{definition}
According to the use of seed alignments $\mathcal{I}$, the EA task is classified into (semi-) supervised and self-supervised or unsupervised settings. The (semi-) supervised setting leverages part of $\mathcal{I}$ as supervision signals for learning, while the self-supervised or unsupervised setting does not require any supervision.

%% file: paragraphs/3.method.tex
\begin{figure*}[t]
\centering
\includegraphics[width=1.0\linewidth]{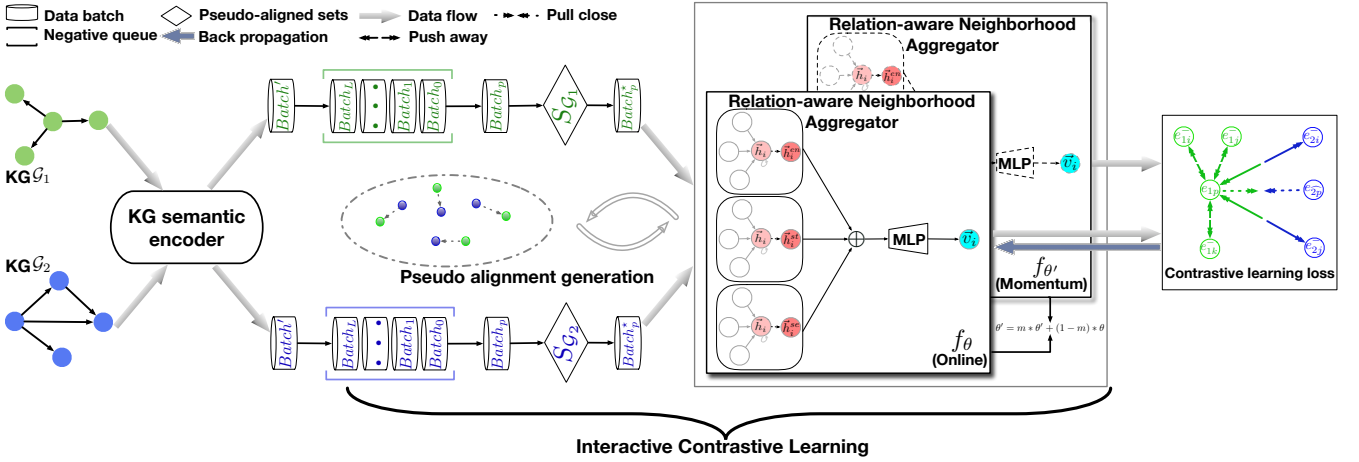}
\caption{The overall framework of \solution. It consists of three main parts: (1) KG semantic encoder; (2) The relation-aware neighborhood aggregator; (3) Interactive contrastive learning mechanism.}
\label{fig:framework}
\end{figure*}

\section{The Proposed Approach}
\label{sec:method}

Although supervised EA methods achieve SOTA performance, their dependence on supervision signals limits real-world EA applications. In contrast, self-supervised methods show the ability to obtain competitive performance without any supervision, and have much better scalability. To fully exploit the potential of self-supervised EA approaches, three crucial issues remain to be addressed.
\begin{itemize}
    \item How to obtain the entity initial representation embedding efficiently and effectively by KG's side information?
    \item How to fully utilize and integrate the structural and semantic KG information in a self-supervised EA framework?
    \item How to establish a direct cross-KG information interaction channel for a self-supervised EA framework instead of only performing unidirectional intra-KG contrastive learning?  
\end{itemize}

To address the above issues, we introduce the interactive contrastive learning for self-supervised EA framework, which contains three parts: \textbf{1. A KG semantic encoder} is utilized to encode the names, descriptions of entities and relations for providing efficient initial entity embeddings. \textbf{2. A relation-aware neighborhood aggregator} is introduced to fully exploit relations' structural and semantic information so as to update entity representation embeddings. \textbf{3. An interactive contrastive learning mechanism} is proposed to construct pseudo-aligned entity pairs as virtual pivots for guiding contrastive learning between two KGs, thus promoting the direct learning of cross-KG interactions. Figure~\ref{fig:framework} shows the overall framework of our proposed approach \solution.

\subsection{KG Semantic Encoder}

KG semantic encoder aims to fully capture the semantics of entities and relations by leveraging their names and descriptions. Since pre-trained language models (PLMs) have achieved remarkable progress in Natural Language Processing (NLP), we decide to leverage LaBSE~\cite{feng2020language}, a language-agnostic pre-trained sentence model, for encoding the name. Besides, we choose SentenceTransformers\footnote{https://github.com/UKPLab/sentence-transformers}, a transformer-based sentence embedding framework, to obtain entity description embeddings.
LaBSE is a SOTA language-agnostic sentence embedding PLM which is trained on 109 different languages. SentenceTransformers is a Python framework for SOTA sentence, text and image embeddings. 
We choose them due to their outstanding empirical performance in capturing precise semantics for phrases and long sentences while helping us to cross the multilingual barrier.

\noindent \textbf{KG Name Encoder.} The names of each entity $e$ and relation $r$ are usually composed of phrases. We utilize the tokenizer of LaBSE to obtain the tokens of the entity name ${n_{e}}$ and relation name ${n_{r}}$, and get corresponding embeddings through LaBSE model $f_{\theta_{LaBSE}}$. Next, a mean pooling operation is applied to the embedding of each token, followed by an $L_2$ normalization.
\begin{equation}
\vec{h}_{n_x}  = \left \| \operatorname{Mean}( f_{\theta_{LaBSE}}(n_x)) \right \|_{L_2}, n_x \in \left \{ n_e, n_r   \right \}.
\label{eq:1}
\end{equation}

\noindent \textbf{Entity Description Encoder.} Moreover, entity description contains rich semantics that allows PLMs to encode with more context information. The entity description is usually composed of one or more sentences and contains multifaceted features related to the entity. For each description, we select the first $L_{max}$ characters to feed into the SentenceTransformers model $f_{\theta_{ST}}$ and get the representation of entity description $d_e$,
\begin{equation}
\vec{h}_{d_e} = \left \| f_{\theta_{ST}}(d_e)  \right \|_{L_2}.
\label{eq:2}
\end{equation}

Finally, we get the entity $e$'s representation embedding $\vec{h}_{e}$ by concatenating entity name embedding $\vec{h}_{n_e}$ and entity description embedding $\vec{h}_{d_e}$, and more processing details are available in Section~\ref{sec:im_detail}, 

\begin{equation}
\vec{h}_{e} = \operatorname{Concat}(\vec{h}_{n_e} , \vec{h}_{d_e}).
\label{eq:3}
\end{equation}

\subsection{Relation-Aware Neighborhood Aggregator}

Relation-aware neighborhood aggregator aims to update entity embeddings by performing message passing with the help of KG's relation information which can aggregate neighborhood entities' helpful information for EA to the center node entity. Refer to Section~\ref{sec:icl}, our self-supervised EA framework has the online and momentum parts as shown in Figure~\ref{fig:framework}. These two parts have the same model structure, and they have different parameter updating strategies. We take the online part as an example to introduce this module. The KG's relations can bring us two aspects of crucial information: \emph{\textbf{structural}} --- neighbor entities, which provide valuable context information for understanding the center node entity, and \emph{\textbf{semantic}} --- neighbor relations, which capture rich semantic information of edges adjacent to the given center node entity. In this work, we apply GAT~\cite{velivckovic2017graph} and its variants as the backbone networks due to its effectiveness.

\noindent \textbf{Structural Aggregator.} The aggregation of neighbor entities is performed by considering both entities' importance and their relations to the center node. One vanilla GAT aggregates all neighbor entity embeddings for the center node to model the importance of different entities, which treats all relations equally, formally,  
\begin{equation}
\begin{split}
\vec{h}_{i}^{en}&=\parallel_{k=1}^{K} \sigma\left(\sum_{j \in \mathcal{N}_{i} \cup \left \{ e_i \right \} } \alpha_{i j}^{k} \mathbf{W}^{k} \vec{h}_{j}\right), \\
\alpha_{i j}&=\frac{\exp \left(\sigma \left(\overrightarrow{\mathbf{q}}^{T}\left[\mathbf{W} \vec{h}_{i} \| \mathbf{W} \vec{h}_{j}\right]\right)\right)}{\sum_{k \in \mathcal{N}_{i} \cup \left \{ e_i \right \} } \exp \left(\sigma \left(\overrightarrow{\mathbf{q}}^{T}\left[\mathbf{W} \vec{h}_{i} \| \mathbf{W} \vec{h}_{k}\right]\right)\right)},
\end{split}
\label{eq:4}
\end{equation}
 where $\mathbf{W}$ is the learnable linear transformation's weight matrix. The attention mechanism is a single-layer feedforward neural network parametrized by a weight vector $\overrightarrow{\mathbf{q}}$. $K$ is the multi-head attention number. $\sigma$ is a nonlinear activation function such as $\operatorname{LeakyReLU}$. $\alpha_{i j}$ is the normalized attention coefficient. $(.)^{T}$ represents matrix transposition and $\|$ is vector concatenation operation.
 
 Another extended GAT is proposed to model the importance of different adjacent relations. It uses relation specific attention heads as relation-wise gates to control information flow from neighbor entities and obtains updated entity embedding $\vec{h}_{i}^{st}$,
\begin{equation}
\small
\begin{split}
\vec{h}_{i}^{st}&=\parallel_{k=1}^{K} \sigma\left(\sum_{j \in \mathcal{N}_{i} } \beta_{i j}^{k} \mathbf{W}^{k} \vec{h}_{j}\right), \\
\beta_{i j}&=\frac{\exp \left(\gamma_{i j}\right)}{\sum_{k \in \mathcal{N}_{i} } \exp \left(\gamma_{i k}\right)},\\
\gamma_{i j}&=\sigma\left(\sigma \left(\vec{r}_{i j} \mathbf{W}_{r_{ij}^{1}}+\mathbf{B}_{r_{ij}^{1}}\right) \mathbf{W}_{r_{ij}^{2}}+\mathbf{B}_{r_{ij}^{2}}\right),
\end{split}
\label{eq:66}
\end{equation}
where $\vec{r}_{i j}$ is trainable relation specific embedding between entity $e_i$ and entity $e_j$. $\mathbf{W}_{r_{ij}}$ and $\mathbf{B}_{r_{ij}}$ are relation's transformation and bias weight matrix.

\noindent \textbf{Semantic Aggregator.} Relation's name label in KGs usually contains certain semantic information. For example, the ``\emph{sisters}''\footnote{https://dbpedia.org/property/sisters} relation in DBpedia describes a kinship relation between the head and tail entities, we argue that information is helpful for obtaining better entity representation embeddings for self-supervised EA. The semantic aggregator aims to aggregate and fuse neighbor relations' semantic information into the center node entity representation embedding. In this work, we apply a one-hop GAT that is the same as Equation~\ref{eq:4} to fuse neighbor relation name label embeddings into corresponding center entity embeddings. Specifically, for each $\vec{h}_j$ in Equation~\ref{eq:4}, we replace it with a average neighbor relation embedding $\widehat{h}_{j} = \frac{1}{K_{ij}} \sum_{x=0}^{K_{ij}} \vec{h}_{n_{r_x}}$, where $n_{r_x}$ is the $x$-th relation's name label between entity $e_i$ and entity $e_j$, $K_{ij}$ is the total number of relations between entity $e_i$ and entity $e_j$, and $\vec{h}_{n_{r_x}}$ is the corresponding relation name label embedding, which can be calculated from Equation~\ref{eq:1}. For $\vec{h}_k$ in Equation~\ref{eq:4}, we replace it with $\widehat{h}_{k}$ using a similar method. This way, we can obtain $\vec{h}_i^{se}$ of entity $e_i$ based on its neighbor relation name labels.

Finally, we use a fully-connected layer to fuse three aspects of embeddings $\vec{h}_{i}^{en}$, $\vec{h}_{i}^{st}$, and $\vec{h}_{i}^{se}$ to obtain the final entity representation embedding $\vec{v}_{i}$,

\begin{equation}
\vec{v}_{i} = \operatorname{MLP} ( \operatorname{Concat} ( \vec{h}_{i}^{en},\vec{h}_{i}^{st},\vec{h}_{i}^{se})).
\label{eq:rel_se_3}
\end{equation}

\subsection{Interactive Contrastive Learning}
\label{sec:icl}

Interactive contrastive learning mechanism is designed to learn direct cross-KG interactions for self-supervised EA. It mainly consists of three parts: \textbf{(1)} Momentum contrastive learning mechanism performs intra-KG contrastive learning for self-supervised EA, which samples negative entities and pushes them far away from the positive one in the same source KG, thus the aligned entity pairs in different KGs are relatively drawn close. \textbf{(2)} Negative sample queues store previously processed batches as negative samples for the positive batch, which can provide a large number of negative samples for the training process. \textbf{(3)} Interactive contrastive learning mechanism constructs pseudo-aligned pairs as virtual pivots during training and establishes cross-KG direct information interaction for self-supervised EA. It pulls the positive samples and pseudo-aligned entities closer, while pushing them far away from the negative samples in different KGs.

\noindent \textbf{Momentum Contrastive Learning Mechanism.} Given a training KG $\mathcal{G}_1$'s initial entity embeddings $\mathbf{H_1}=\left\{\vec{h}_{1}, \vec{h}_{2}, \ldots, \vec{h}_{{\left | \mathcal{E}_1 \right |}}\right\}$ which can be calculated from Equation~\ref{eq:3}, self-supervised EA representation learning aims to learn an online embedding tranformation function $f_{\theta}$ that maps $\mathbf{H_1}$ to $\mathbf{V_1}=\left\{\vec{v}_{1}, \vec{v}_{2}, \ldots, \vec{v}_{{\left | \mathcal{E}_1 \right |}}\right\}$ with $\vec{v}_{x} = f_{\theta} \left ( \vec{h}_{x}  \right ) $, such that $\vec{v}_{x}$ best describes $\vec{h}_{x}$. Instance-wise contrastive learning achieves this objective by optimizing a contrastive loss. Following SelfKG~\cite{liu2021self}, we use the Noise Contrastive Estimation (NCE) loss\cite{gutmann2010noise}. In practice, we jointly optimize the NCE loss on both source KG $\mathcal{G}_1$ and target KG $\mathcal{G}_2$, taking $\mathcal{G}_1$ as example, the NCE loss can be defined as,

\begin{equation}
\small
\begin{split}
&\mathcal{L}_{\text{nce}_{\mathcal{G}_1}} = \sum_{x=1}^{|\mathcal{E}_1|}-\log \frac{\exp \left(\vec{v}_{x} \cdot \vec{v}_{x}^{\prime} / \tau\right)}{\exp \left(\vec{v}_{x} \cdot \vec{v}_{x}^{\prime} / \tau\right)+\sum_{k=1}^{r} \exp \left(\vec{v}_{x} \cdot \vec{v}_{k}^{\prime} / \tau\right)} \\
&= \sum_{x=1}^{|\mathcal{E}_1|} \underbrace{-\frac{1}{\tau}\vec{v}_{x} \cdot \vec{v}_{x}^{\prime}}_{\rm Alignment} + \underbrace{\log ( \exp \left(\vec{v}_{x} \cdot \vec{v}_{x}^{\prime} / \tau\right)+\sum_{k=1}^{r} \exp \left(\vec{v}_{x} \cdot \vec{v}_{k}^{\prime} / \tau\right) )}_{\rm Uniformity}  ,
\end{split}
\label{eq:nce}
\end{equation}
where, the ``Alignment'' part pulls the positive pair closer while the ``Uniformity'' part pushes the negative pairs away, $\vec{v}_{x}$ and $\vec{v}_{x}^{\prime}$ are positive embeddings for entity $e_{x}$ which can be obtained from the online encoder $f_{\theta}$ and momentum encoder $f_{\theta^{\prime}}$, and $\vec{v}_{k}^{\prime}$ includes $r$ negative samples' embeddings, and $\tau$ is a temperature hyperparameter. The online encoder's parameter $\theta$ is instantly updated with the backpropagation. The negative sample's embedding is obtained by feeding corresponding $\vec{h}_{x}$ to the momentum encoder parameterized by $\theta^{\prime}$, $\vec{v}_{x}^{\prime} = f_{\theta^{\prime}}(\vec{h}_{x})$, where $\theta^{\prime}$ is a moving average of $\theta$,
\begin{equation}
\theta^{\prime} \leftarrow m \times \theta^{\prime}+(1-m) \times \theta, m \in[0,1).
\label{eq:momentum_update}
\end{equation}

\noindent \textbf{Negative Sample Queues.} While performing contrastive learning for self-supervised EA, we need to sample a large number of negative entities from the same source KG to avoid the conflict by simply excluding the positive one. We maintain two negative queues for both KGs that store previously processed batches as negative samples. As shown in the middle part of Figure~\ref{fig:framework}, when a new data batch $Batch^{\prime}$ arrives, we add it to the corresponding negative queue tail, and the head $Batch_{0}$ in the same negative queue is dequeued as a positive sample batch $Batch_{p}$. In the early stage of training, we do not perform any parameter update until one of the negative queues reaches the predefined length $L + 1$, where $L$ for the number of previous processed batches used as negative samples and ``$1$'' for the dequeued positive batch. Let the numbers of the entities in KGs $|\mathcal{E}_1|$, $|\mathcal{E}_2|$, the batch size $B$ and $L$ is constraint by,
\begin{equation}
(L+1) \times B \leqslant  \min \left(\left|\mathcal{E}_{1}\right|,\left|\mathcal{E}_{2}\right|\right).
\label{eq:number}
\end{equation}
Finally, the real number of negative samples used for each positive batch's entity is $(L + 1) \times B - 1$.

\newcommand\mycommfont[1]{\tiny\ttfamily\textcolor{PineGreen}{#1}}
\SetCommentSty{mycommfont}
\begin{algorithm}[h]
	\DontPrintSemicolon
	\SetNoFillComment
	\small
	\textbf{Input:} Online encoder $f_\theta$, momentum encoder $f_{\theta^{\prime}}$, training datasets $X_1,X_2$ of KGs $\mathcal{G}_1,\mathcal{G}_2$, distance threshold $\lambda$, momentum $m$.\\	
    $\theta^{\prime}=\theta$.   \tcp*{initialize momentum encoder as the online encoder}
    $\mathbf{H_1},\mathbf{H_2}=\operatorname{KG\_Semantic\_Encoder}(X_1,X_2)$ \tcp*{get the initial entity embeddings for source KG $\mathcal{G}_1$ and target KG $\mathcal{G}_2$}
    $NQ_1,NQ_2 = \emptyset$ \tcp*{initialize negative sample queues for $\mathcal{G}_1$ and $\mathcal{G}_2$}
	\While{$\mathrm{not~MaxEpoch}$}   
	{
	\tcc{Get pseudo-aligned entities}
	$\mathbf{V_1}=f_{\theta}(\mathbf{H_1})$,
	$\mathbf{V_2}=f_{\theta}(\mathbf{H_2})$ \tcp*{get features for all data}
	$Pair_{1 \rightarrow 2}=Faiss.search(\mathbf{V_1}, \mathbf{V_2}, 1, \lambda)$ \\
	$Pair_{2 \rightarrow 1}=Faiss.search(\mathbf{V_2}, \mathbf{V_1}, 1, \lambda)$ \\
	\tcp*{Find the top 1 nearest neighbor with L2 distance less than $\lambda$ for each entity}
	$Pair_{all}=\operatorname{Merge}(Pair_{1 \rightarrow 2},Pair_{2  \rightarrow 1})$
	
	\tcc{Training}
	\For ( \tcp*[f]{load a minibatch $Batch^{\prime}$}) {$Batch^{\prime}~\mathbf{in}~\mathrm{Dataloader}(X_1 + X_2)$}	
	{
	    \eIf{$Batch^{\prime}$ belongs to $X_1$}
	    {
	    $NQ_1$.add\_tail($Batch^{\prime}$), $NQ = NQ_1$, $\mathcal{G} = \mathcal{G}_1$
	    }{
	    $NQ_2$.add\_tail($Batch^{\prime}$), $NQ = NQ_2$, $\mathcal{G} = \mathcal{G}_2$
	    }
	    \If{$NQ$.Length == $L+1$}
	    {
	    $Batch_{p}$ = $NQ$.dequeue\_head$()$ \tcp*{the head batch of corresponding negative sample queue dequeues as the positive batch}
		$\vec{v}_{x}=f_\theta(Batch_{p}),\vec{v}_{x}^{\prime}=f_{\theta^{\prime}}(Batch_{p})$\\ $\vec{v}_{x}^{\prime\prime}=f_{\theta^{\prime}}(Pair_{all}[Batch_{p}])$ \tcp*{forward pass through the online encoder and momentum encoder, $Batch_{p}$ as positive samples}
		$\mathbf{NV}_{\mathcal{G}_1} = f_{\theta^{\prime}}(NQ_1)$, $\mathbf{NV}_{\mathcal{G}_2} =f_{\theta^{\prime}}(NQ_2)$  \\
		$Loss =\mathcal{L}_\mathrm{\text{nce}_{\mathcal{G}}}(\vec{v}_{x},\vec{v}_{x}^{\prime},\mathbf{NV}_{\mathcal{G}})+ \mathcal{L}_\mathrm{\text{icl}_{\mathcal{G}}}(\vec{v}_{x},\vec{v}_{x}^{\prime\prime},\mathbf{NV}_{\mathcal{G}_1},\mathbf{NV}_{\mathcal{G}_2})$ 
		\tcp*{calculate loss with Equations~\ref{eq:nce} and ~\ref{eq:icl}}
        $\theta=\mathrm{Adam}(Loss,\theta)$ \tcp*{update online encoder's parameters}
		$\theta^{\prime} = $m$ \times \theta^{\prime} + (1-$m$) \times \theta$
		\tcp*{update momentum encoder's parameters}
		}
	}    	
}
	\caption{\small Interactive Contrastive Learning Process.}
\label{alg:icl}
\end{algorithm}

\noindent \textbf{Interactive Contrastive Learning Mechanism.} The interactive contrastive learning mechanism aims to build a direct information interaction channel for two KGs during training. We construct pseudo-aligned entity sets $S_{\mathcal{G}_1}$ and $S_{\mathcal{G}_2}$ for each source KG to the corresponding target KG at the beginning of each training epoch. Given initial embeddings of source and target KGs $\mathcal{G}_1$ and $\mathcal{G}_2$, $\mathbf{H_1}$ and $\mathbf{H_2}$, we feed them into the online encoder $f_{\theta}$ to obtain the corresponding entity embeddings $\mathbf{V_1}$ and $\mathbf{V_2}$. For each entity $e_{1p}$ in $\mathcal{G}_1$ and $e_{2q}$ in $\mathcal{G}_2$, we match the most similar entities ${e_{\widehat{2p}}}$ and ${e_{\widehat{1q}}}$ from the corresponding $\mathcal{G}_2$ and $\mathcal{G}_1$. We predefine an $L_2$ distance threshold $\lambda$, if $\operatorname{Dis}(e_{1p},{e_{\widehat{2p}}})$ or $\operatorname{Dis}(e_{2q},{e_{\widehat{1q}}})$ is less than $\lambda$, we add the pair of entities $\left \langle e_{1p},{e_{\widehat{2p}}} \right \rangle$ or $\left \langle e_{2q},{e_{\widehat{1q}}} \right \rangle$ to the corresponding pseudo-aligned set $S_{\mathcal{G}_1}$ or $S_{\mathcal{G}_2}$. We use these pseudo-aligned entity pairs as virtual pivots to guide the entire contrastive learning process, thus establishing the cross-KG direct information interaction for self-supervised EA. We apply Faiss\footnote{https://github.com/facebookresearch/faiss} for obtaining pseudo-aligned sets efficiently. It is worth noting that distinct from the previous iterative bootstrapping strategy~\cite{sun2018bootstrapping} that maintains high-confidence aligned entity pairs iteratively, our approach focuses on generating larger pseudo-aligned sets automatically, which liberates us from the caution of introducing additional noises. Although the pseudo-aligned sets we construct also contain some noises, our experimental results in the next section demonstrate that our model is less sensitive to those noises, and the advantages of the pseudo-aligned sets on self-supervised EA far outweigh the disadvantages of noises. To incorporate pseudo-aligned entity pairs into the contrastive learning process we introduce an additional NCE loss $\mathcal{L}_{\text{icl}}$ into the training process, and in Equation~\ref{eq:icl} we take the example of the KG $\mathcal{G}_1$, where $\vec{v}_{x} = f_{\theta}(\vec{h}_{1p}), \vec{v}_{x}^{\prime}=f_{\theta^{\prime}}(\vec{h}_{\widehat{2p}})$, $\vec{h}_{1p}$ and $\vec{h}_{\widehat{2p}}$ are the embeddings of the corresponding positve entity and its pseudo-aligned entity. The hyperparameter $\beta$ balances the sensitiveness of the model to the negative sample $\vec{v}_{1_k}^{\prime}$ and $\vec{v}_{2_k}^{\prime}$ from different KGs, 
\begin{equation}
\small
\begin{split}
\mathcal{L}_{\text{icl}_{\mathcal{G}_1}} = \sum_{x=1}^{|\mathcal{E}_1|} \beta& \cdot \left [ -\log \frac{\exp \left(\vec{v}_{x} \cdot \vec{v}_{x}^{\prime} / \tau\right)}{\exp \left(\vec{v}_{x} \cdot \vec{v}_{x}^{\prime} / \tau\right)+\sum_{k=1}^{r} \exp \left(\vec{v}_{x} \cdot \vec{v}_{1_k}^{\prime} / \tau\right)} \right ] \\
+ (1 - \beta)& \cdot \left [ -\log \frac{\exp \left(\vec{v}_{x} \cdot \vec{v}_{x}^{\prime} / \tau\right)}{\exp \left(\vec{v}_{x} \cdot \vec{v}_{x}^{\prime} / \tau\right)+\sum_{k=1}^{r} \exp \left(\vec{v}_{x} \cdot \vec{v}_{2_k}^{\prime} / \tau\right)} \right ].
\end{split}
\label{eq:icl}
\end{equation}

Based on the above discussion, we can get the overall optimization goal $\mathcal{L}$ of \solution in Equation~\ref{eq:icl_all}. A pseudo-code of our proposed algorithm is given in Algorithm~\ref{alg:icl}, 
\begin{equation}
\mathcal{L}=\sum\limits_{i \in \{ 1,2 \}}^{} \mathcal{L}_{\text{nce}_{\mathcal{G}_i}}+\mathcal{L}_{\text{icl}_{\mathcal{G}_i}}.
\label{eq:icl_all}
\end{equation}

%% file: paragraphs/4.experiments.tex
\section{Experiments}
\label{sec:experiment}

In this section, we evaluate our proposed approach on DBP15K, a widely used benchmark for EA. We first introduce the experimental settings, then report the overall results, and finally conduct ablation studies as well as parameter sensitivity analyses.
\subsection{Experimental Settings}
\label{sec:im_detail}
\noindent\textbf{Dataset.} The DBP15K dataset is originally built by~\cite{JAPE}, which includes three cross-lingual datasets extracted from DBpedia\footnote{\url{http://downloads.dbpedia.org/2016-04/}}. Each contains $15,000$ reference alignments between English (EN) and one of the other languages, i.e., Chinese (ZH), Japanese (JA) and French (FR). Table~\ref{table:dbp15k} presents the detailed statistics about DBP15K.
To make wide comparisons, we report the results on both orignal~\cite{JAPE} and translated version~\cite{xu2019cross-lingual}, corresponding to two different experimental settings: multilingual setting and monolingual setting. In the multilingual setting, all the used data are extracted from original multilingual DBpedia without using any translation tools.  
Moreover, to make the comparisons fair, following previous works~\cite{xu2019cross-lingual,wang2020knowledge}, we use Google Translate to translate all non-English entity descriptions into English when using translated DBP15K in the monolingual setting.

\begin{table}[!htp]
	\centering
	\scriptsize
	\caption{Statistics of the simplified DBP15K datasets.}
	\begin{tabular}{l|l|ccccc}
		\toprule[1.6pt]
		\multicolumn{2}{c|}{\textbf{Datasets}} & \textbf{Ent.} & \textbf{Rel.} & \textbf{R-Tri.} & \textbf{Ent Alignments.} \\
		\midrule
		\multirow{2}{*}{ZH\_EN} & ZH & 19,388 & 1,700 & 70,414 & 15,000 \\
		& EN & 19,572 & 1,322 & 95,142 & 15,000 \\
		\midrule
		\multirow{2}{*}{JA\_EN} & JA & 19,814 & 1,298 & 77,214 & 15,000 \\
		& EN & 19,780 & 1,152 & 93,484 & 15,000\\
		\midrule
		\multirow{2}{*}{FR\_EN} & FR & 19,661 & 902 & 105,998 & 15,000 \\
		& EN & 19,993 & 1,207 & 115,722 & 15,000\\
		\bottomrule[1.6pt]
	\end{tabular}

	\label{table:dbp15k}
\end{table}

\begin{table}[h]
	\centering
    \caption{Overall results on DBP15K. \textmd{Methods marked with ``$^*$'' use a translated version of DBP15K in the monolingual setting. The best results in supervised/unsupervised or self-supervised settings are marked in underline/bold.}}
    \renewcommand\tabcolsep{3.5pt}
	\input{table/dbp15k}

    \label{table:overall_performace}
\end{table}


\noindent \textbf{Evaluation Metrics.} We use \textbf{Hits@N} as the evaluation metric. \textbf{Hits@N} means the proportion of correct entities that rank no larger than N (\textbf{N} is $1$ and $10$), and higher Hits@N indicates better performance. We further calculate the average Hits@1 of three subtasks to measure the overall performance.

\noindent \textbf{Baselines.} We compare our \textbf{ICLEA} against two groups of baselines, i.e., semi-supervised and un/self-supervised EA approaches. The latter include MultiKE~\cite{zhang2019multi}, EVA~\cite{liu2021visual_eva} and SelfKG~\cite{liu2020self}, and the former are further divided into three branches: \textbf{(1)} KG structure embedding-based methods that only use the KGs' structure information: MTransE~\cite{MTransE}, JAPE~\cite{JAPE}, IPTransE~\cite{zhu2017iterative}, SEA~\cite{pei2019semi}, KECG~\cite{li2019semi}, MuGNN~\cite{MuGNN}, RSNs~\cite{guo2019learning}, AliNet~\cite{sun2020knowledge}, BootEA~\cite{sun2018bootstrapping}, NAEA~\cite{zhu2019neighborhood}, MRPEA~\cite{shi2019modeling} and PREA~\cite{mao2020relational}. \textbf{(2)} Graph-based methods that leverage GNNs to utilize various KG information: GM-Align\cite{xu2019cross-lingual}, RDGCN~\cite{wu2019relation}, HGCN\cite{wu2019jointly}, AttrGNN~\cite{liu2020exploring}, DGMC~\cite{fey2020deep}, RNM~\cite{zhu2020relation}, EPEA~\cite{wang2020knowledge} and CEAFF~\cite{CEAFF}. \textbf{(3)} BERT-based methods that directly use multilingual BERT to deal with EA: HMAN~\cite{yang2019aligning} and BERT-INT~\cite{tang2019bert-int}.

\noindent \textbf{Data Preprocessing.} The name labels of entity and relation in DBP15K usually have a string with meaningless prefixes, such as ``http://dbpedia.org/resource/Jay\_Chou'', we remove the meaningless prefixes and replace the underscores with empty spaces, producing the final entity and relation name labels. \cite{tang2019bert-int} provides the original multilingual description data of entities in DBP15K, we directly use these entity description data in the multilingual setting. In the monolingual setting, we leverage an open source Google Translate tool~\footnote{https://github.com/ssut/py-googletrans} to translate the non-English entity descriptions into the corresponding English versions. To obtain one-hop neighbors for each entity, we treat KGs as undirected graphs, that means we use the relation triples in the datasets to find all entities connected to a given entity, regardless of the connection's direction.

\noindent \textbf{Model and Training Details.} Our model is implemented with Pytorch $1.7.0$. We employ Adam as our optimizer with a learning rate $1\mathrm{e}{-6}$ and gradually reduce it to maintain the stability of training. The number of training epochs is $300$, the batch size is $64$, momentum $m$ is $0.9999$, similarity threshold $\lambda$ is $1.0$, temperature $\tau$ is $0.08$ and the size of negative queue is $32$, the entity description length is $512$. We use one-hop GAT in our model for its efficiency, during our experiments we find that using multi-hop GAT brings some performance degradation, which may be caused by the KGs' heterogeneity and the amplified neighbor noises in the self-supervised EA setting, that is consistent with the findings of previous work~\cite{liu2020self}. We utilize the data about a center entity and its up to 15 neighboring entities as the input of GAT. In the relation-aware neighborhood aggregator layer, the dimension of both input and output embeddings is equal to $LaBSE\_DIM + DESC\_DIM$, and in the fully-connected layer, the input dimension is $LaBSE\_DIM \times 7 $ and the output dimension is $LaBSE\_DIM \times 5$, where $LaBSE\_DIM$ and $DESC\_DIM$ are equal to 768. We choose $L_2$ distance as entity embedding distance metric for pseudo alignment generation and set $\beta$ to $0.9$ in interactive contrastive learning. We randomly select $5\%$ from the original training set as the validation set in order to early stopping. All experiments are conducted on a Ubuntu server with eight GPUs (GeForceRTX 3090), each evaluation is repeated five times with the same random seed (set to 37) and the averaged results are reported.

\begin{figure*}[t]
\centering
\includegraphics[width=1.0\linewidth]{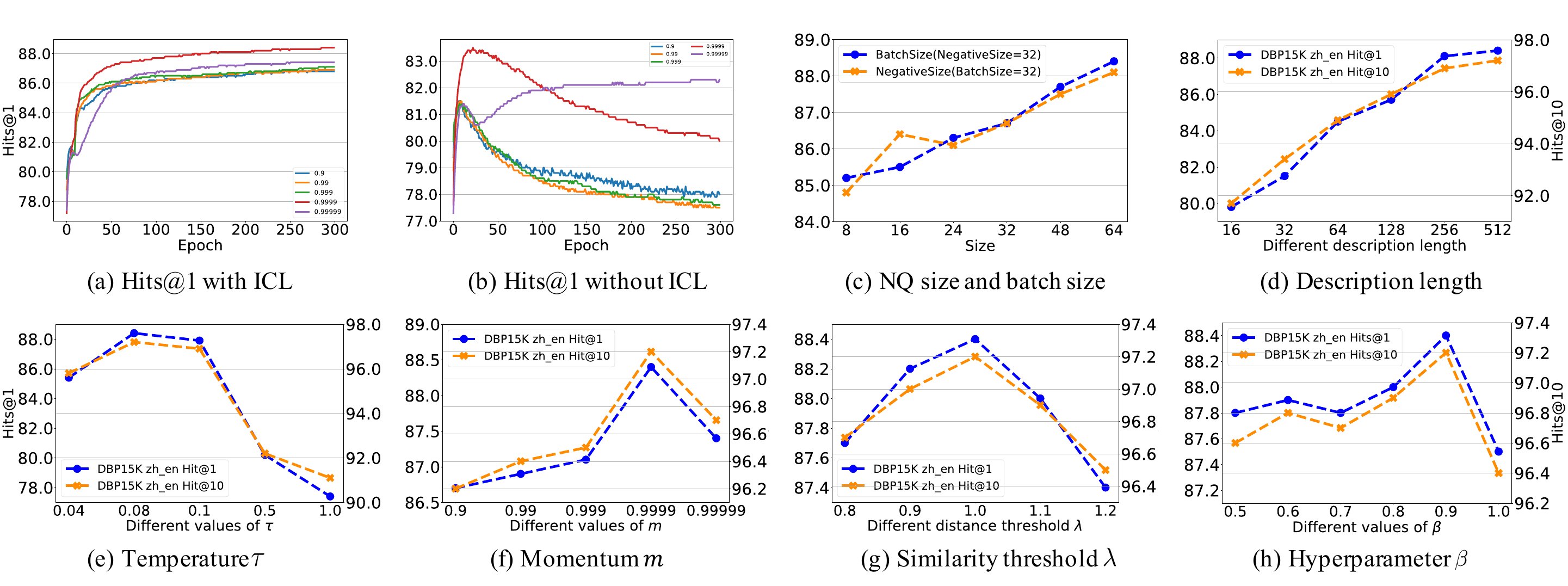}
\caption{Study on negative queue (NQ) size, batch size, description length, temperature $\tau$, momentum $m$, $L_2$ distance threshold $\lambda$ and hyperparameter $\beta$ on DBP15K$_{\text{ZH\_EN}}$. (a), (b) present the test Hits@1 curve throughout the training with and without ICL.}
\label{fig:all_ab_study}
\end{figure*}

\begin{table*}[!htp]
\scriptsize
\centering
\caption{Ablation study of \solution on DBP15K. \textmd{``$^*$'' means using translated version of DBP15K in the monolingual setting.}}
\scalebox{1.12}{
      \begin{tabular}{@{}cccccccccccccc@{}}
          \toprule[1.6pt]
          \multirow{2}{*}{Model} &
            \multicolumn{2}{|c|}{DBP15K$_{\text{ZH\_EN}}$} &
            \multicolumn{2}{|c}{DBP15K$_{\text{JA\_EN}}$} &
            \multicolumn{2}{|c}{DBP15K$_{\text{FR\_EN}}$} &
            \multicolumn{2}{|c}{DBP15K$_{\text{ZH\_EN}}$$^*$} &
            \multicolumn{2}{|c}{DBP15K$_{\text{JA\_EN}}$$^*$} &
            \multicolumn{2}{|c}{DBP15K$_{\text{FR\_EN}}$$^*$} &
            \multirow{2}{*}{\makecell[c]{}} \\ 
            \cmidrule(l){2-13} 
          &
            \multicolumn{1}{|c}{Hits@1} &
            \multicolumn{1}{c|}{Hits@10} &
            \multicolumn{1}{c}{Hits@1} &
            \multicolumn{1}{c|}{Hits@10} &
            \multicolumn{1}{c}{Hits@1} &
            \multicolumn{1}{c|}{Hits@10} &
            \multicolumn{1}{c}{Hits@1} &
            \multicolumn{1}{c|}{Hits@10} &
            \multicolumn{1}{c}{Hits@1} &
            \multicolumn{1}{c|}{Hits@10} &
            \multicolumn{1}{c}{Hits@1} &
            \multicolumn{1}{c}{Hits@10} &\\ 
            \midrule
         \multicolumn{1}{c|}{\textbf{\solution}} &
            \multicolumn{1}{c}{\textbf{88.4}} &
            \multicolumn{1}{c|}{\textbf{97.2}} &
            \multicolumn{1}{c}{\textbf{91.9}} &
            \multicolumn{1}{c|}{\textbf{97.5}} &
            \multicolumn{1}{c}{\textbf{98.6}} &
            \multicolumn{1}{c|}{\textbf{99.9}} &
            \multicolumn{1}{c}{\textbf{92.1}} &
            \multicolumn{1}{c|}{\textbf{98.1}} &
            \multicolumn{1}{c}{\textbf{95.5}} &
            \multicolumn{1}{c|}{\textbf{98.8}} &
            \multicolumn{1}{c}{\textbf{99.2}} &
            \multicolumn{1}{c}{\textbf{99.9}}             \\  

        \multicolumn{1}{c|}{\textbf{w/o Nam.}} &
            \multicolumn{1}{c}{72.0} &
            \multicolumn{1}{c|}{92.7} &
            \multicolumn{1}{c}{74.1} &
            \multicolumn{1}{c|}{94.1} &
            \multicolumn{1}{c}{89.5} &
            \multicolumn{1}{c|}{99.3} &
            \multicolumn{1}{c}{75.8} &
            \multicolumn{1}{c|}{94.0} &
            \multicolumn{1}{c}{82.0} &
            \multicolumn{1}{c|}{97.0} &
            \multicolumn{1}{c}{89.2} &
            \multicolumn{1}{c}{99.1} \\
            
        \multicolumn{1}{c|}{\textbf{w/o Des.}} &
            \multicolumn{1}{c}{80.4} &
            \multicolumn{1}{c|}{91.4} &
            \multicolumn{1}{c}{87.3} &
            \multicolumn{1}{c|}{93.1} &
            \multicolumn{1}{c}{97.3} &
            \multicolumn{1}{c|}{99.5} &
            \multicolumn{1}{c}{86.5} &
            \multicolumn{1}{c|}{93.4} &
            \multicolumn{1}{c}{92.4} &
            \multicolumn{1}{c|}{96.6} &
            \multicolumn{1}{c}{97.1} &
            \multicolumn{1}{c}{99.4} \\
            
        \multicolumn{1}{c|}{\textbf{w/o ICL}} &
            \multicolumn{1}{c}{84.3} &
            \multicolumn{1}{c|}{94.1} &
            \multicolumn{1}{c}{88.8} &
            \multicolumn{1}{c|}{96.2} &
            \multicolumn{1}{c}{97.1} &
            \multicolumn{1}{c|}{99.8} &  
            \multicolumn{1}{c}{89.3} &
            \multicolumn{1}{c|}{96.3} &
            \multicolumn{1}{c}{93.3} &
            \multicolumn{1}{c|}{98.1} &
            \multicolumn{1}{c}{97.3} &
            \multicolumn{1}{c}{99.8} \\
            
        \multicolumn{1}{c|}{\textbf{w/o MCL}} &
            \multicolumn{1}{c}{86.7} &
            \multicolumn{1}{c|}{96.5} &
            \multicolumn{1}{c}{90.2} &
            \multicolumn{1}{c|}{97.1} &
            \multicolumn{1}{c}{97.6} &
            \multicolumn{1}{c|}{99.7} &
            \multicolumn{1}{c}{91.4} &
            \multicolumn{1}{c|}{97.9} &
            \multicolumn{1}{c}{94.2} &
            \multicolumn{1}{c|}{98.3} &
            \multicolumn{1}{c}{97.9} &
            \multicolumn{1}{c}{99.8} \\

        \multicolumn{1}{c|}{\textbf{w/o Rel.}} &
            \multicolumn{1}{c}{87.0} &
            \multicolumn{1}{c|}{96.7} &
            \multicolumn{1}{c}{90.6} &
            \multicolumn{1}{c|}{97.2} &
            \multicolumn{1}{c}{97.8} &
            \multicolumn{1}{c|}{99.9} &
            \multicolumn{1}{c}{90.5} &
            \multicolumn{1}{c|}{97.3} &
            \multicolumn{1}{c}{94.9} &
            \multicolumn{1}{c|}{98.5} &
            \multicolumn{1}{c}{98.1} &
            \multicolumn{1}{c}{99.8} \\            

          \bottomrule[1.6pt]
      \end{tabular}

  }

\label{table:ab_study}
\end{table*}

\subsection{Main Results}

Table~\ref{table:overall_performace} lists the overall performance. For all baselines, we take the reported results from the original papers and ~\cite{liu2021self}. From the results, we have the following key observations:

\noindent \textbf{Comparisons with semi-supervised methods.} \solution outperforms almost all previous supervised models on both original and translated DBP15K, and achieves results comparable to BERT-INT (the strongest baseline). They use multilingual BERT directly to help with entity alignment. \solution achieves the same excellent results as BERT-INT on the FR\_EN subtask, where the BERT-INT's Hits@1 is 0.6\% higher than that of \solution, while the \solution's Hits@10 is 0.1\% higher than that of BERT-INT. The gap between \solution and BERT-INT is only 4.5\% in average Hits@1, and it is further narrowed to 1.9\% on the translated dataset (though BERT-INT does not translate dataset directly, it uses multilingual BERT). It is worth mentioning that BERT-INT actually fine-tunes the whole multilingual BERT, whose parameter scale is ten times larger than \solution (110M vs. 11M). In a nutshell, \solution significantly narrows the gap between supervised and self-supervised EA approaches.

\noindent \textbf{Comparisons with un/self-supervised methods.} In the unsupervised or self-supervised setting, \solution notably outperforms all those baselines with a large margin. Even compared with the strongest baseline SelfKG, \solution improves the average Hits@1 score by 9.0\% on the original datasets in the multilingual setting and 6.3\% on the translated datasets in the monolingual setting. Note that the improvement is more significant in original datasets in the multilingual setting, especially in ZH\_EN subtask, indicating that \solution is less language dependent and suitable for dealing with the cross-lingual entity alignment tasks.

\noindent \textbf{Multilingual bias phenomenon.} \solution exhibits multilingual bias consistent with most previous methods, i.e., performing best on the FR\_EN subtask and worst on the ZH\_EN subtask. Early models that rely on only KG's structure information to handle EA e.g. MTransE, JAPE, IPTransE, SEA, KECG, etc., do not exhibit such multilingual bias phenomenon. This may be due to the fact that as EA evolves, more and more methods incorporate entity-related semantic information in the model to assist EA, while leading to the multilingual bias phenomenon described earlier. Note that EVA has a relatively balanced performance on three datasets due to the introduction of language-independent entity images. This inspires us that multi-modal entity information utilization in self-supervised EA would be a promising future direction.

\subsection{Ablation Study}

We perform an ablation study to evaluate the effectiveness of all model components. Accordingly, we implement five variants of \solution by removing entity name information (\textit{w/o Nam.}), entity description information (\textit{w/o Des.}), relation-aware neighborhood aggregator (\textit{w/o Rel.}), momentum contrastive learning mechanism (\textit{w/o MCL}) and interactive contrastive learning mechanism (\textit{w/o ICL}). Table~\ref{table:ab_study} presents the results, which show that the removal of each component has a negative impact on the performance. From the results, we have the following key observations:

\noindent \textbf{The entity name information has the greatest impact on the model's accuracy.} Specifically, \textit{w/o Nam.} has the largest negative impact on Hits@1 value, demonstrating the entity name information plays a fundamental role in the model's accuracy. Intuitively, the entity description information contains richer semantic information than entity name labels and should play a more critical role in the EA model's accuracy. However, the experimental results are inconsistent with the intuition, and we believe this may be due to the heterogeneity of entity descriptions in different KGs. There may be differences in the focus, order, etc. of different KGs for the same entity's description, so that noise is introduced along with entity semantics, which may be the reason for this phenomenon. We believe how to solve the entity description's heterogeneity problem and better incorporate description information in self-supervised EA would be a valuable open future problem to explore.

\noindent \textbf{The entity description information has the greatest impact on the model's Hits@10.} As shown in Table~\ref{table:ab_study}, in most cases, \textit{w/o Des.} has the largest negative impact on the Hits@10 value, demonstrating the effectiveness of entity description. The Hits@10 value responds to a certain extent to the recall ability of the model, which indicates that entity description information plays a crucial role in improving the model's recall.

\noindent \textbf{The relation information can also improve the model's accuracy.} \textit{w/o Rel.} causes relatively more significant decreases in Hits@1 than Hits@10, indicating the importance of relation and structure in distinguishing more ambiguous cases.


\noindent \textbf{In addition to the high-performance improvement, the interactive contrastive learning mechanism also brings more stable model training process.} \textit{w/o ICL} brings a more notable performance degradation, illustrating the importance of the interactive contrastive learning mechanism for self-supervised EA. Besides, we further investigate how the performance changes during the model training epochs with and without ICL. As shown in Figure~\ref{fig:all_ab_study}a and~\ref{fig:all_ab_study}b, without ICL, the test Hits@1 score usually reaches the peak after tens of epochs and then shows a sharp drop, while using ICL brings a more stable model training process.

\begin{figure}[t]
\centering
\includegraphics[width=0.72\linewidth]{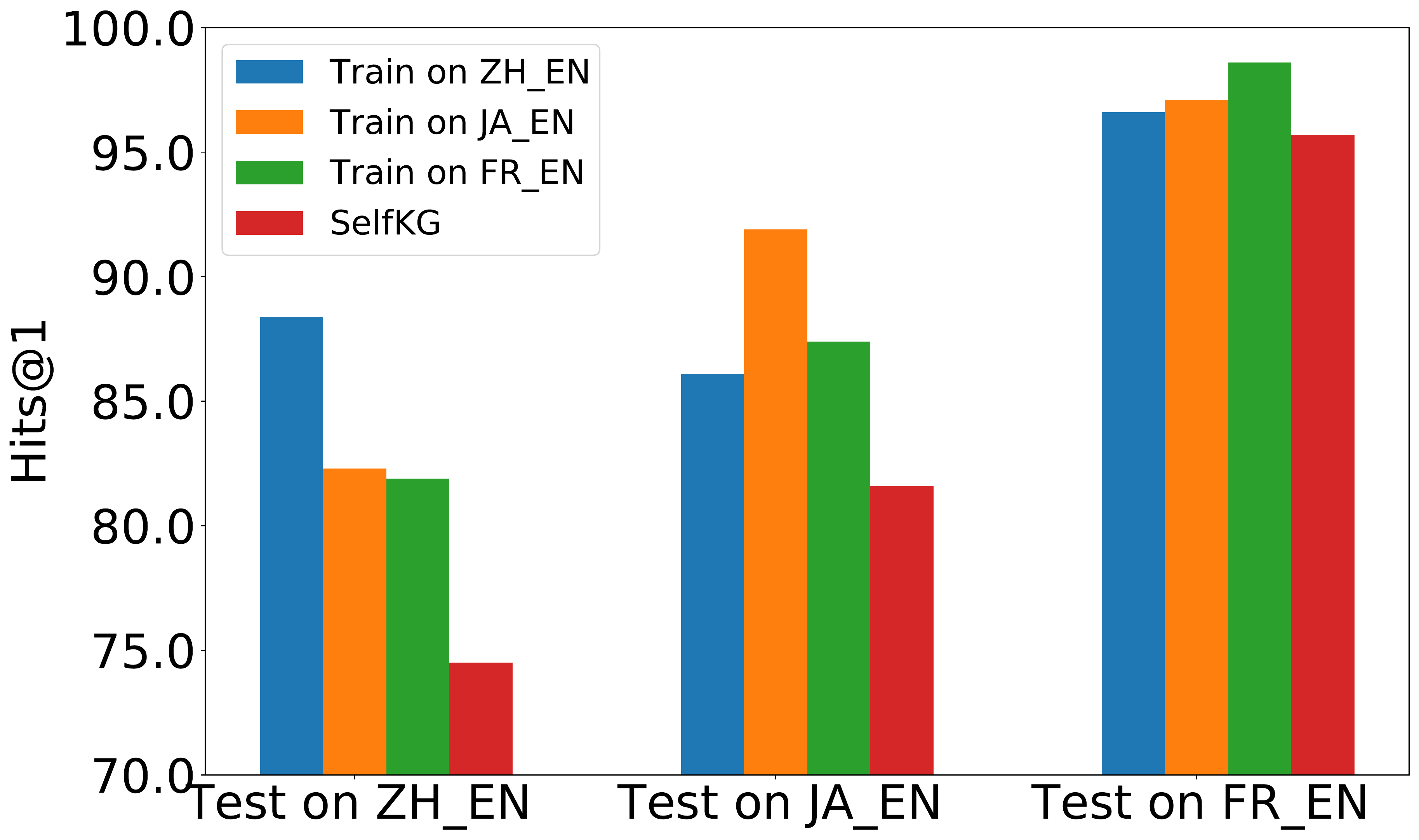}
\caption {Inductive learning of \solution in multilingual setting.}
\label{fig:inductive}
\end{figure}

\subsection{Hyper-parameters Analyses}

\noindent \textbf{Negative queue size and batch size.} For these two parameters, we perform a grid search from 8 to 64. Figure~\ref{fig:all_ab_study}c shows that the performance exhibits a fluctuating upward trend when fixing batch size to 64 and increasing the size of the negative queue. Meanwhile, Figure~\ref{fig:all_ab_study}c also shows the performance increases steadily with larger batch size when setting queue size to 64. While both expanding negative queue and batch size can improve the performance, a larger batch size usually brings more computational cost. 

\noindent \textbf{Entity description length.} As shown in Figure~\ref{fig:all_ab_study}d, a longer description brings better performance because it can provide more semantic information. According to our statistical results, the description length of most entities in original DBP15K is less than 512. Considering the computation cost and the description length distribution in the datasets, we set the maximal length to 512.

\noindent \textbf{Temperature $\tau$ and momentum coefficient $m$.} The temperature hyperparameter $\tau$ regulates the degree of attention to difficult samples~\cite{wang2021understanding} and the momentum coefficient $m$ prevents model's sensitive update~\cite{he2020momentum}. We empirically choose both parameters from finite sets and present the results in Figure~\ref{fig:all_ab_study}e and Figure~\ref{fig:all_ab_study}f, which show that a relatively larger $m$ (e.g., 0.9999) leads to better performance and $\tau=0.08$ is a good cut-off point.

\noindent \textbf{Distance threshold $\lambda$ and hyperparameter $\beta$.} Distance threshold $\lambda$ in Equation~\ref{eq:icl} controls the selection of pseudo alignments. A larger $\lambda$ means more pseudo alignments for cross-KG interaction, while possibly introducing more noises. We find in our experiments that our model is less sensitive to these noises, and a proper proportion of noise can even bring some help to the model's performance. As shown in Figure~\ref{fig:all_ab_study}g, it is a good balance point when $\lambda = 1.0$. It will be an interesting research problem to design a dynamic adjustment mechanism for $\lambda$ to further balance the effect of noise on the model. The hyperparameter $\beta$ is the trade-off factor between the positive and negative samples from different KGs. As shown in Figure~\ref{fig:all_ab_study}h, we find that the sensitivity of the model is more preferred to the source KG's negative sample queue, and thus set $\beta$ to $0.9$.

\subsection{Inductive Learning}

We train \solution on one dataset of original DBP15K, then directly transfer and evaluate them on the other two datasets which can evaluate our model's cross-lingual transference ability, and show the results in Figure~\ref{fig:inductive}. When training \solution on FR\_EN and testing on ZH\_EN/JA\_EN, the Hits@1 values show a significant decrease, with relative reductions of 7.4\% and 4.9\%, indicating that the cross-lingual transference from French to Chinese and Japanese is extremely challenging for the model. When training \solution on ZH\_EN/JA\_EN and testing on JA\_EN/ZH\_EN, the Hits@1 values decrease by 6.3\% and 6.9\%, respectively, which reveals Chinese and Japanese are almost the same difficult to be transferred between each other. No matter training on any dataset, the test performance on FR\_EN is comparable to the normal one (i.e., train and test on FR\_EN), this is very likely due to the fact that French and English are more similar to each other, making the alignment avoids the affect of different languages. We also find that no matter training on ZH\_EN, JA\_EN, or FR\_EN, \solution's test performance on the other two datasets outperforms SelfKG by a large margin (Hits@1 values increased by at least 9.9\%, 5.5\%, and 3.1\%, respectively), which demonstrates \solution is better at overcoming the multilingual challenge in self-supervised entity alignment.

%% file: table/dbp15k.tex
\scalebox{0.8}{
    \setlength{\belowcaptionskip}{-0.1cm}
      \begin{tabular}{@{}cccccccc@{}}
          \toprule[2pt]
          \multirow{2}{*}{Model} &
            \multicolumn{2}{|c|}{DBP15K$_{\text{ZH\_EN}}$} &
            \multicolumn{2}{|c}{DBP15K$_{\text{JA\_EN}}$} &
            \multicolumn{2}{|c|}{DBP15K$_{\text{FR\_EN}}$} &
            \multirow{2}{*}{\makecell[c]{AVG\\Hits@1}} \\ 
            \cmidrule(l){2-7} 
           &
            \multicolumn{1}{|c}{Hits@1} &
            \multicolumn{1}{c|}{Hits@10} &
            \multicolumn{1}{c}{Hits@1} &
            \multicolumn{1}{c|}{Hits@10} &
            \multicolumn{1}{c}{Hits@1} &
            \multicolumn{1}{c|}{Hits@10} &\\ 
            \midrule
          \multicolumn{8}{c}{Supervised}                     \\ 
          \midrule
          \multicolumn{1}{c|}{\begin{tabular}[c]{@{}c@{}}MTransE \end{tabular}} &
            \multicolumn{1}{c}{30.8} &
            \multicolumn{1}{c|}{61.4} &
            \multicolumn{1}{c}{27.9} &
            \multicolumn{1}{c|}{57.5} &
            \multicolumn{1}{c}{24.4} &
            \multicolumn{1}{c|}{55.6} &
            27.7\\ 
          \multicolumn{1}{c|}{JAPE} &
            \multicolumn{1}{c}{41.2} &
            \multicolumn{1}{c|}{74.5} &
            \multicolumn{1}{c}{36.3} &
            \multicolumn{1}{c|}{68.5} &
            \multicolumn{1}{c}{32.4} &
            \multicolumn{1}{c|}{66.7} &
            36.6\\
          \multicolumn{1}{c|}{IPTransE} &
            \multicolumn{1}{c}{40.6} &
            \multicolumn{1}{c|}{73.5} &
            \multicolumn{1}{c}{36.7} &
            \multicolumn{1}{c|}{69.3} &
            \multicolumn{1}{c}{33.3} &
            \multicolumn{1}{c|}{68.5} &
            36.9\\ 
          \multicolumn{1}{c|}{SEA} &
            \multicolumn{1}{c}{42.4} &
            \multicolumn{1}{c|}{79.6} &
            \multicolumn{1}{c}{38.5} &
            \multicolumn{1}{c|}{78.3} &
            \multicolumn{1}{c}{40.0} &
            \multicolumn{1}{c|}{79.7} &
            40.3\\
          \multicolumn{1}{c|}{KECG} &
            \multicolumn{1}{c}{47.8} &
            \multicolumn{1}{c|}{83.5} &
            \multicolumn{1}{c}{49.0} &
            \multicolumn{1}{c|}{84.4} &
            \multicolumn{1}{c}{48.6} &
            \multicolumn{1}{c|}{85.1} &
            48.5\\ 
          \multicolumn{1}{c|}{MuGNN} &
            \multicolumn{1}{c}{49.4} &
            \multicolumn{1}{c|}{84.4} &
            \multicolumn{1}{c}{50.1} &
            \multicolumn{1}{c|}{85.7} &
            \multicolumn{1}{c}{49.5} &
            \multicolumn{1}{c|}{87.0} &
            49.7\\ 
          \multicolumn{1}{c|}{RSNs} &
            \multicolumn{1}{c}{50.8} &
            \multicolumn{1}{c|}{74.5} &
            \multicolumn{1}{c}{50.7} &
            \multicolumn{1}{c|}{73.7} &
            \multicolumn{1}{c}{51.6} &
            \multicolumn{1}{c|}{76.8} &
            51.0\\ 
          \multicolumn{1}{c|}{AliNet} &
            \multicolumn{1}{c}{53.9} &
            \multicolumn{1}{c|}{82.6} &
            \multicolumn{1}{c}{54.9} &
            \multicolumn{1}{c|}{83.1} &
            \multicolumn{1}{c}{55.2} &
            \multicolumn{1}{c|}{85.2} &
            54.7\\ 
          \multicolumn{1}{c|}{BootEA} &
            \multicolumn{1}{c}{62.9} &
            \multicolumn{1}{c|}{84.8} &
            \multicolumn{1}{c}{62.2} &
            \multicolumn{1}{c|}{85.4} &
            \multicolumn{1}{c}{65.3} &
            \multicolumn{1}{c|}{87.4} &
            63.5\\ 
          \multicolumn{1}{c|}{NAEA} &
            \multicolumn{1}{c}{65.0} &
            \multicolumn{1}{c|}{86.7} &
            \multicolumn{1}{c}{64.1} &
            \multicolumn{1}{c|}{87.3} &
            \multicolumn{1}{c}{67.3} &
            \multicolumn{1}{c|}{89.4} &
            65.5\\ 
          \multicolumn{1}{c|}{MRPEA} &
            \multicolumn{1}{c}{68.1} &
            \multicolumn{1}{c|}{86.7} &
            \multicolumn{1}{c}{65.5} &
            \multicolumn{1}{c|}{85.9} &
            \multicolumn{1}{c}{67.7} &
            \multicolumn{1}{c|}{89.0} &
            67.1\\ 
          \multicolumn{1}{c|}{PREA} &
            \multicolumn{1}{c}{71.5} &
            \multicolumn{1}{c|}{92.9} &
            \multicolumn{1}{c}{71.3} &
            \multicolumn{1}{c|}{93.3} &
            \multicolumn{1}{c}{73.9} &
            \multicolumn{1}{c|}{94.6} &
            72.2\\ 
            \midrule
          \multicolumn{1}{c|}{GM-Align$^*$} &
            \multicolumn{1}{c}{67.9} &
            \multicolumn{1}{c|}{78.5} &
            \multicolumn{1}{c}{74.0} &
            \multicolumn{1}{c|}{87.2} &
            \multicolumn{1}{c}{89.4} &
            \multicolumn{1}{c|}{95.2} &
            77.1\\
          \multicolumn{1}{c|}{{RDGCN$^*$}} &
            \multicolumn{1}{c}{70.8} &
            \multicolumn{1}{c|}{84.6} &
            \multicolumn{1}{c}{76.7} &
            \multicolumn{1}{c|}{89.5} &
            \multicolumn{1}{c}{88.6} &
            \multicolumn{1}{c|}{95.7} &
            78.7\\
          \multicolumn{1}{c|}{{HGCN$^*$}} &
            \multicolumn{1}{c}{72.0} &
            \multicolumn{1}{c|}{85.7} &
            \multicolumn{1}{c}{76.6} &
            \multicolumn{1}{c|}{89.7} &
            \multicolumn{1}{c}{89.2} &
            \multicolumn{1}{c}{96.1} &
            79.3\\
          \multicolumn{1}{c|}{AttrGNN$^*$} &
            \multicolumn{1}{c}{79.6} &
            \multicolumn{1}{c|}{92.9} &
            \multicolumn{1}{c}{78.3} &
            \multicolumn{1}{c|}{92.1} &
            \multicolumn{1}{c}{91.9} &
            \multicolumn{1}{c|}{97.8} &
            83.3\\
          \multicolumn{1}{c|}{{DGMC$^*$}} &
            \multicolumn{1}{c}{80.1} &
            \multicolumn{1}{c|}{87.5} &
            \multicolumn{1}{c}{84.8} &
            \multicolumn{1}{c|}{89.7} &
            \multicolumn{1}{c}{93.3} &
            \multicolumn{1}{c|}{96.0} &
            86.1\\
           \multicolumn{1}{c|}{RNM$^*$} &
             \multicolumn{1}{c}{{84.0}} &
             \multicolumn{1}{c|}{{91.9}} &
             \multicolumn{1}{c}{{87.2}} &
             \multicolumn{1}{c|}{{94.4}} &
             \multicolumn{1}{c}{{93.8}} &
             \multicolumn{1}{c|}{{95.4}} &
             88.3\\
          \multicolumn{1}{c|}{EPEA$^*$} &
             \multicolumn{1}{c}{{88.5}} &
             \multicolumn{1}{c|}{{95.3}} &
             \multicolumn{1}{c}{{92.4}} &
             \multicolumn{1}{c|}{{96.9}} &
             \multicolumn{1}{c}{{95.5}} &
             \multicolumn{1}{c|}{{98.6}} &
             92.1\\
          \multicolumn{1}{c|}{CEAFF} &
            \multicolumn{1}{c}{79.5} &
            \multicolumn{1}{c|}{-} &
            \multicolumn{1}{c}{86.0} &
            \multicolumn{1}{c|}{-} &
            \multicolumn{1}{c}{96.4} &
            \multicolumn{1}{c|}{-} &
            87.3\\ 
            \midrule
          \multicolumn{1}{c|}{HMAN} &
            \multicolumn{1}{c}{87.1} &
            \multicolumn{1}{c|}{98.7} &
            \multicolumn{1}{c}{93.5} &
            \multicolumn{1}{c|}{99.4} &
            \multicolumn{1}{c}{97.3} &
            \multicolumn{1}{c|}{\underline{99.8}} &
            92.6\\ 
          \multicolumn{1}{c|}{BERT-INT} &
            \multicolumn{1}{c}{\underline{96.8}} &
            \multicolumn{1}{c|}{\underline{99.0}} &
            \multicolumn{1}{c}{\underline{96.4}} &
            \multicolumn{1}{c|}{\underline{99.1}} &
            \multicolumn{1}{c}{\underline{99.2}} &
            \multicolumn{1}{c|}{\underline{99.8}} &
            97.5\\ 
          \midrule
            \multicolumn{8}{c}{Unsupervised or Self-supervised}              \\ 
            \midrule
          \multicolumn{1}{c|}{MultiKE} &
            \multicolumn{1}{c}{50.9} &
            \multicolumn{1}{c|}{57.6} &
            \multicolumn{1}{c}{39.3} &
            \multicolumn{1}{c|}{48.9} &
            \multicolumn{1}{c}{63.9} &
            \multicolumn{1}{c|}{71.2} &
            51.4\\
          \multicolumn{1}{c|}{EVA} &
            \multicolumn{1}{c}{75.2} &
            \multicolumn{1}{c|}{89.5} &
            \multicolumn{1}{c}{73.7} &
            \multicolumn{1}{c|}{89.0} &
            \multicolumn{1}{c}{73.1} &
            \multicolumn{1}{c|}{90.9} &
            74.0\\            
         \multicolumn{1}{c|}{SelfKG} &
            \multicolumn{1}{c}{74.5} &
            \multicolumn{1}{c|}{86.6} &
            \multicolumn{1}{c}{81.6} &
            \multicolumn{1}{c|}{91.3} &
            \multicolumn{1}{c}{95.7} &
            \multicolumn{1}{c|}{99.2} &
            84.0\\ 
         \multicolumn{1}{c|}{SelfKG$^*$} &
            \multicolumn{1}{c}{82.9} &
            \multicolumn{1}{c|}{91.9} &
            \multicolumn{1}{c}{89.0} &
            \multicolumn{1}{c|}{95.3} &
            \multicolumn{1}{c}{95.9} &
            \multicolumn{1}{c|}{99.2} &
            89.3\\ 
             \midrule[1.3pt]
           \multicolumn{1}{c|}{\textbf{\solution}} &
            \multicolumn{1}{c}{\textbf{88.4}} &
            \multicolumn{1}{c|}{\textbf{97.2}} &
            \multicolumn{1}{c}{\textbf{91.9}} &
            \multicolumn{1}{c|}{\textbf{97.5}} &
            \multicolumn{1}{c}{\textbf{98.6}} &
            \multicolumn{1}{c|}{\textbf{99.9}} &
            \textbf{93.0}\\ 
            
            \multicolumn{1}{c|}{{\solution}$^*$} &
            \multicolumn{1}{c}{92.1} &
            \multicolumn{1}{c|}{98.1} &
            \multicolumn{1}{c}{95.5} &
            \multicolumn{1}{c|}{98.8} &
            \multicolumn{1}{c}{99.2} &
            \multicolumn{1}{c|}{99.9} &
            95.6\\ 
          \bottomrule[2pt]
      \end{tabular}
  }

%% file: paragraphs/5.conclusion.tex
\section{Conclusion and Future Work}
\label{sec:conlusion}
In this work, we propose a model named \solution for self-supervised EA. To better utilize the entity's structural and semantic information, we separately encode entity names and descriptions with the help of PLMs and propose a relation-aware neighborhood aggregator to better leverage the structural and semantic information brought by KGs' relations. We design an innovative interactive contrastive learning mechanism by constructing pseudo-aligned entity pairs to establish a direct information interaction channel for the two KGs. Experimental results show that \solution performs on par with previous SOTA supervised counterparts and outperforms previous best self-supervised results by a large margin. We also present two promising directions for self-supervised EA. (1) Aligned entities usually have similar attribute-value pairs in different KGs. How to combine entity attributes into a self-supervised EA framework will be a focus of our future work. (2) PCL~\cite{li2020prototypical} introduces prototypes as latent variables to help find the maximum-likelihood estimation of model's parameters in an Expectation-Maximization framework. KG entities also have potential prototype information, such as people, buildings, etc. How to let the model learn this information and assist in self-supervised EA is also an open challenge.

%% file: paragraphs/7.acknowledgement.tex
\section{Acknowledgement}
This work is supported by the National Key Research and Development Program of China (2020AAA0106501), the NSFC Youth Project (61825602), grants from Joint Institute of Tsinghua University-China Mobile Communications Group Co.,Ltd., BAAI and the Institute for Guo Qiang, Tsinghua University (2019GQB0003).

%% file: sigir22-sigconf.bbl

\begin{thebibliography}{63}


\ifx \showCODEN    \undefined \def \showCODEN     #1{\unskip}     \fi
\ifx \showDOI      \undefined \def \showDOI       #1{#1}\fi
\ifx \showISBNx    \undefined \def \showISBNx     #1{\unskip}     \fi
\ifx \showISBNxiii \undefined \def \showISBNxiii  #1{\unskip}     \fi
\ifx \showISSN     \undefined \def \showISSN      #1{\unskip}     \fi
\ifx \showLCCN     \undefined \def \showLCCN      #1{\unskip}     \fi
\ifx \shownote     \undefined \def \shownote      #1{#1}          \fi
\ifx \showarticletitle \undefined \def \showarticletitle #1{#1}   \fi
\ifx \showURL      \undefined \def \showURL       {\relax}        \fi
\providecommand\bibfield[2]{#2}
\providecommand\bibinfo[2]{#2}
\providecommand\natexlab[1]{#1}
\providecommand\showeprint[2][]{arXiv:#2}

\bibitem[Bruna et~al\mbox{.}(2014)]%
        {bruna2013spectral}
\bibfield{author}{\bibinfo{person}{Joan Bruna}, \bibinfo{person}{Wojciech
  Zaremba}, \bibinfo{person}{Arthur Szlam}, {and} \bibinfo{person}{Yann
  LeCun}.} \bibinfo{year}{2014}\natexlab{}.
\newblock \showarticletitle{Spectral networks and deep locally connected
  networks on graphs}. In \bibinfo{booktitle}{\emph{2nd International
  Conference on Learning Representations, ICLR 2014}}.
\newblock


\bibitem[Cao et~al\mbox{.}(2019a)]%
        {MuGNN}
\bibfield{author}{\bibinfo{person}{Yixin Cao}, \bibinfo{person}{Zhiyuan Liu},
  \bibinfo{person}{Chengjiang Li}, \bibinfo{person}{Juanzi Li}, {and}
  \bibinfo{person}{Tat-Seng Chua}.} \bibinfo{year}{2019}\natexlab{a}.
\newblock \showarticletitle{Multi-Channel Graph Neural Network for Entity
  Alignment}. In \bibinfo{booktitle}{\emph{Proceedings of the 57th Annual
  Meeting of the Association for Computational Linguistics}}.
  \bibinfo{pages}{1452--1461}.
\newblock


\bibitem[Cao et~al\mbox{.}(2019b)]%
        {cao2019unifying}
\bibfield{author}{\bibinfo{person}{Yixin Cao}, \bibinfo{person}{Xiang Wang},
  \bibinfo{person}{Xiangnan He}, \bibinfo{person}{Zikun Hu}, {and}
  \bibinfo{person}{Tat-Seng Chua}.} \bibinfo{year}{2019}\natexlab{b}.
\newblock \showarticletitle{Unifying knowledge graph learning and
  recommendation: Towards a better understanding of user preferences}. In
  \bibinfo{booktitle}{\emph{The world wide web conference}}.
  \bibinfo{pages}{151--161}.
\newblock


\bibitem[Chen et~al\mbox{.}(2020b)]%
        {chen2020jointly_sigir}
\bibfield{author}{\bibinfo{person}{Chong Chen}, \bibinfo{person}{Min Zhang},
  \bibinfo{person}{Weizhi Ma}, \bibinfo{person}{Yiqun Liu}, {and}
  \bibinfo{person}{Shaoping Ma}.} \bibinfo{year}{2020}\natexlab{b}.
\newblock \showarticletitle{Jointly non-sampling learning for knowledge graph
  enhanced recommendation}. In \bibinfo{booktitle}{\emph{Proceedings of the
  43rd International ACM SIGIR Conference on Research and Development in
  Information Retrieval}}. \bibinfo{pages}{189--198}.
\newblock


\bibitem[Chen et~al\mbox{.}(2017)]%
        {MTransE}
\bibfield{author}{\bibinfo{person}{Muhao Chen}, \bibinfo{person}{Yingtao Tian},
  \bibinfo{person}{Mohan Yang}, {and} \bibinfo{person}{Carlo Zaniolo}.}
  \bibinfo{year}{2017}\natexlab{}.
\newblock \showarticletitle{Multilingual Knowledge Graph Embeddings for
  Cross-lingual Knowledge Alignment}.
\newblock
\urldef\tempurl%
\url{https://doi.org/10.24963/ijcai.2017/209}
\showDOI{\tempurl}


\bibitem[Chen et~al\mbox{.}(2020a)]%
        {chen2020simple}
\bibfield{author}{\bibinfo{person}{Ting Chen}, \bibinfo{person}{Simon
  Kornblith}, \bibinfo{person}{Mohammad Norouzi}, {and}
  \bibinfo{person}{Geoffrey Hinton}.} \bibinfo{year}{2020}\natexlab{a}.
\newblock \showarticletitle{A simple framework for contrastive learning of
  visual representations}. In \bibinfo{booktitle}{\emph{ICML}}. PMLR,
  \bibinfo{pages}{1597--1607}.
\newblock


\bibitem[Cui et~al\mbox{.}(2017)]%
        {z11_cui2019kbqa}
\bibfield{author}{\bibinfo{person}{Wanyun Cui}, \bibinfo{person}{Yanghua Xiao},
  \bibinfo{person}{Haixun Wang}, \bibinfo{person}{Yangqiu Song},
  \bibinfo{person}{Seung-won Hwang}, {and} \bibinfo{person}{Wei Wang}.}
  \bibinfo{year}{2017}\natexlab{}.
\newblock \showarticletitle{KBQA: Learning Question Answering over QA Corpora
  and Knowledge Bases}.
\newblock \bibinfo{journal}{\emph{Proceedings of the VLDB Endowment}}
  \bibinfo{volume}{10}, \bibinfo{number}{5} (\bibinfo{year}{2017}).
\newblock


\bibitem[Defferrard et~al\mbox{.}(2016)]%
        {defferrard2016convolutional}
\bibfield{author}{\bibinfo{person}{Micha{\"e}l Defferrard},
  \bibinfo{person}{Xavier Bresson}, {and} \bibinfo{person}{Pierre
  Vandergheynst}.} \bibinfo{year}{2016}\natexlab{}.
\newblock \showarticletitle{Convolutional neural networks on graphs with fast
  localized spectral filtering}.
\newblock \bibinfo{journal}{\emph{Advances in neural information processing
  systems}}  \bibinfo{volume}{29} (\bibinfo{year}{2016}),
  \bibinfo{pages}{3844--3852}.
\newblock


\bibitem[Feng et~al\mbox{.}(2022)]%
        {feng2020language}
\bibfield{author}{\bibinfo{person}{Fangxiaoyu Feng}, \bibinfo{person}{Yinfei
  Yang}, \bibinfo{person}{Daniel Cer}, \bibinfo{person}{Naveen Arivazhagan},
  {and} \bibinfo{person}{Wei Wang}.} \bibinfo{year}{2022}\natexlab{}.
\newblock \showarticletitle{Language-agnostic BERT Sentence Embedding}. In
  \bibinfo{booktitle}{\emph{Proceedings of the 60th Annual Meeting of the
  Association for Computational Linguistics (Volume 1: Long Papers)}}.
  \bibinfo{pages}{878--891}.
\newblock


\bibitem[Fey et~al\mbox{.}(2020)]%
        {fey2020deep}
\bibfield{author}{\bibinfo{person}{Matthias Fey}, \bibinfo{person}{Jan~E
  Lenssen}, \bibinfo{person}{Christopher Morris}, \bibinfo{person}{Jonathan
  Masci}, {and} \bibinfo{person}{Nils~M Kriege}.}
  \bibinfo{year}{2020}\natexlab{}.
\newblock \showarticletitle{Deep graph matching consensus}.
\newblock \bibinfo{journal}{\emph{arXiv preprint arXiv:2001.09621}}
  (\bibinfo{year}{2020}).
\newblock


\bibitem[Fu et~al\mbox{.}(2020)]%
        {fu2020fairness_sigir}
\bibfield{author}{\bibinfo{person}{Zuohui Fu}, \bibinfo{person}{Yikun Xian},
  \bibinfo{person}{Ruoyuan Gao}, \bibinfo{person}{Jieyu Zhao},
  \bibinfo{person}{Qiaoying Huang}, \bibinfo{person}{Yingqiang Ge},
  \bibinfo{person}{Shuyuan Xu}, \bibinfo{person}{Shijie Geng},
  \bibinfo{person}{Chirag Shah}, \bibinfo{person}{Yongfeng Zhang},
  {et~al\mbox{.}}} \bibinfo{year}{2020}\natexlab{}.
\newblock \showarticletitle{Fairness-aware explainable recommendation over
  knowledge graphs}. In \bibinfo{booktitle}{\emph{Proceedings of the 43rd
  International ACM SIGIR Conference on Research and Development in Information
  Retrieval}}. \bibinfo{pages}{69--78}.
\newblock


\bibitem[Guo et~al\mbox{.}(2019)]%
        {guo2019learning}
\bibfield{author}{\bibinfo{person}{Lingbing Guo}, \bibinfo{person}{Zequn Sun},
  {and} \bibinfo{person}{Wei Hu}.} \bibinfo{year}{2019}\natexlab{}.
\newblock \showarticletitle{Learning to exploit long-term relational
  dependencies in knowledge graphs}. In \bibinfo{booktitle}{\emph{ICML}}. PMLR,
  \bibinfo{pages}{2505--2514}.
\newblock


\bibitem[Gutmann and Hyv{\"a}rinen(2010)]%
        {gutmann2010noise}
\bibfield{author}{\bibinfo{person}{Michael Gutmann} {and} \bibinfo{person}{Aapo
  Hyv{\"a}rinen}.} \bibinfo{year}{2010}\natexlab{}.
\newblock \showarticletitle{Noise-contrastive estimation: A new estimation
  principle for unnormalized statistical models}. In
  \bibinfo{booktitle}{\emph{Proceedings of the Thirteenth International
  Conference on Artificial Intelligence and Statistics}}. JMLR Workshop and
  Conference Proceedings, \bibinfo{pages}{297--304}.
\newblock


\bibitem[Han et~al\mbox{.}(2018)]%
        {z23_han2018neural}
\bibfield{author}{\bibinfo{person}{Xu Han}, \bibinfo{person}{Zhiyuan Liu},
  {and} \bibinfo{person}{Maosong Sun}.} \bibinfo{year}{2018}\natexlab{}.
\newblock \showarticletitle{Neural knowledge acquisition via mutual attention
  between knowledge graph and text}. In \bibinfo{booktitle}{\emph{AAAI}}.
\newblock


\bibitem[Han and Zhao(2009)]%
        {han2009named}
\bibfield{author}{\bibinfo{person}{Xianpei Han} {and} \bibinfo{person}{Jun
  Zhao}.} \bibinfo{year}{2009}\natexlab{}.
\newblock \showarticletitle{Named entity disambiguation by leveraging wikipedia
  semantic knowledge}. In \bibinfo{booktitle}{\emph{Proceedings of the 18th ACM
  conference on Information and knowledge management}}.
  \bibinfo{pages}{215--224}.
\newblock


\bibitem[He et~al\mbox{.}(2020)]%
        {he2020momentum}
\bibfield{author}{\bibinfo{person}{Kaiming He}, \bibinfo{person}{Haoqi Fan},
  \bibinfo{person}{Yuxin Wu}, \bibinfo{person}{Saining Xie}, {and}
  \bibinfo{person}{Ross Girshick}.} \bibinfo{year}{2020}\natexlab{}.
\newblock \showarticletitle{Momentum contrast for unsupervised visual
  representation learning}. In \bibinfo{booktitle}{\emph{CVPR}}.
  \bibinfo{pages}{9729--9738}.
\newblock


\bibitem[Jing and Tian(2020)]%
        {jing2020self}
\bibfield{author}{\bibinfo{person}{Longlong Jing} {and} \bibinfo{person}{Yingli
  Tian}.} \bibinfo{year}{2020}\natexlab{}.
\newblock \showarticletitle{Self-supervised visual feature learning with deep
  neural networks: A survey}.
\newblock \bibinfo{journal}{\emph{IEEE transactions on pattern analysis and
  machine intelligence}} (\bibinfo{year}{2020}).
\newblock


\bibitem[Kenton and Toutanova(2019)]%
        {devlin2018bert}
\bibfield{author}{\bibinfo{person}{Jacob Devlin Ming-Wei~Chang Kenton} {and}
  \bibinfo{person}{Lee~Kristina Toutanova}.} \bibinfo{year}{2019}\natexlab{}.
\newblock \showarticletitle{Bert: Pre-training of deep bidirectional
  transformers for language understanding}. In
  \bibinfo{booktitle}{\emph{Proceedings of NAACL-HLT}}.
  \bibinfo{pages}{4171--4186}.
\newblock


\bibitem[Lehmann et~al\mbox{.}(2015)]%
        {z3_lehmann2015dbpedia}
\bibfield{author}{\bibinfo{person}{Jens Lehmann}, \bibinfo{person}{Robert
  Isele}, {and} \bibinfo{person}{Jakob et al.}}
  \bibinfo{year}{2015}\natexlab{}.
\newblock \showarticletitle{DBpedia--a large-scale, multilingual knowledge base
  extracted from Wikipedia}.
\newblock \bibinfo{journal}{\emph{Semantic web}} \bibinfo{volume}{6},
  \bibinfo{number}{2} (\bibinfo{year}{2015}), \bibinfo{pages}{167--195}.
\newblock


\bibitem[Li et~al\mbox{.}(2019)]%
        {li2019semi}
\bibfield{author}{\bibinfo{person}{Chengjiang Li}, \bibinfo{person}{Yixin Cao},
  \bibinfo{person}{Lei Hou}, \bibinfo{person}{Jiaxin Shi},
  \bibinfo{person}{Juanzi Li}, {and} \bibinfo{person}{Tat-Seng Chua}.}
  \bibinfo{year}{2019}\natexlab{}.
\newblock \showarticletitle{Semi-supervised entity alignment via joint
  knowledge embedding model and cross-graph model}. In
  \bibinfo{booktitle}{\emph{EMNLP}}. \bibinfo{pages}{2723--2732}.
\newblock


\bibitem[Li et~al\mbox{.}(2008)]%
        {li2008rimom}
\bibfield{author}{\bibinfo{person}{Juanzi Li}, \bibinfo{person}{Jie Tang},
  \bibinfo{person}{Yi Li}, {and} \bibinfo{person}{Qiong Luo}.}
  \bibinfo{year}{2008}\natexlab{}.
\newblock \showarticletitle{Rimom: A dynamic multistrategy ontology alignment
  framework}.
\newblock \bibinfo{journal}{\emph{TKDE}} \bibinfo{volume}{21},
  \bibinfo{number}{8} (\bibinfo{year}{2008}), \bibinfo{pages}{1218--1232}.
\newblock


\bibitem[Li et~al\mbox{.}(2020)]%
        {li2020prototypical}
\bibfield{author}{\bibinfo{person}{Junnan Li}, \bibinfo{person}{Pan Zhou},
  \bibinfo{person}{Caiming Xiong}, {and} \bibinfo{person}{Steven Hoi}.}
  \bibinfo{year}{2020}\natexlab{}.
\newblock \showarticletitle{Prototypical Contrastive Learning of Unsupervised
  Representations}. In \bibinfo{booktitle}{\emph{International Conference on
  Learning Representations}}.
\newblock


\bibitem[Liu et~al\mbox{.}(2021b)]%
        {liu2021activeea}
\bibfield{author}{\bibinfo{person}{Bing Liu}, \bibinfo{person}{Harrisen
  Scells}, \bibinfo{person}{Guido Zuccon}, \bibinfo{person}{Wen Hua}, {and}
  \bibinfo{person}{Genghong Zhao}.} \bibinfo{year}{2021}\natexlab{b}.
\newblock \showarticletitle{ActiveEA: Active Learning for Neural Entity
  Alignment}. In \bibinfo{booktitle}{\emph{Proceedings of the 2021 Conference
  on Empirical Methods in Natural Language Processing}}.
  \bibinfo{pages}{3364--3374}.
\newblock


\bibitem[Liu et~al\mbox{.}(2021a)]%
        {liu2021visual_eva}
\bibfield{author}{\bibinfo{person}{Fangyu Liu}, \bibinfo{person}{Muhao Chen},
  \bibinfo{person}{Dan Roth}, {and} \bibinfo{person}{Nigel Collier}.}
  \bibinfo{year}{2021}\natexlab{a}.
\newblock \showarticletitle{Visual Pivoting for (Unsupervised) Entity
  Alignment}. In \bibinfo{booktitle}{\emph{Proceedings of the AAAI Conference
  on Artificial Intelligence (AAAI)}}.
\newblock


\bibitem[Liu et~al\mbox{.}(2022)]%
        {liu2021self}
\bibfield{author}{\bibinfo{person}{Xiao Liu}, \bibinfo{person}{Haoyun Hong},
  \bibinfo{person}{Xinghao Wang}, \bibinfo{person}{Zeyi Chen},
  \bibinfo{person}{Evgeny Kharlamov}, \bibinfo{person}{Yuxiao Dong}, {and}
  \bibinfo{person}{Jie Tang}.} \bibinfo{year}{2022}\natexlab{}.
\newblock \showarticletitle{SelfKG: Self-Supervised Entity Alignment in
  Knowledge Graphs}. In \bibinfo{booktitle}{\emph{Proceedings of the ACM Web
  Conference 2022}} (Virtual Event, Lyon, France) \emph{(\bibinfo{series}{WWW
  '22})}. \bibinfo{publisher}{Association for Computing Machinery},
  \bibinfo{address}{New York, NY, USA}, \bibinfo{pages}{860–870}.
\newblock
\showISBNx{9781450390965}
\urldef\tempurl%
\url{https://doi.org/10.1145/3485447.3511945}
\showDOI{\tempurl}


\bibitem[Liu et~al\mbox{.}(2021c)]%
        {liu2020self}
\bibfield{author}{\bibinfo{person}{Xiao Liu}, \bibinfo{person}{Fanjin Zhang},
  \bibinfo{person}{Zhenyu Hou}, \bibinfo{person}{Li Mian},
  \bibinfo{person}{Zhaoyu Wang}, \bibinfo{person}{Jing Zhang}, {and}
  \bibinfo{person}{Jie Tang}.} \bibinfo{year}{2021}\natexlab{c}.
\newblock \showarticletitle{Self-supervised learning: Generative or
  contrastive}.
\newblock \bibinfo{journal}{\emph{IEEE Transactions on Knowledge and Data
  Engineering}} (\bibinfo{year}{2021}).
\newblock


\bibitem[Liu et~al\mbox{.}(2020)]%
        {liu2020exploring}
\bibfield{author}{\bibinfo{person}{Zhiyuan Liu}, \bibinfo{person}{Yixin Cao},
  \bibinfo{person}{Liangming Pan}, \bibinfo{person}{Juanzi Li}, {and}
  \bibinfo{person}{Tat-Seng Chua}.} \bibinfo{year}{2020}\natexlab{}.
\newblock \showarticletitle{Exploring and Evaluating Attributes, Values, and
  Structure for Entity Alignment}. In \bibinfo{booktitle}{\emph{Proceedings of
  the 2020 Conference on Empirical Methods in Natural Language Processing
  (EMNLP)}}. \bibinfo{pages}{6355--6364}.
\newblock


\bibitem[Mahdisoltani et~al\mbox{.}(2014)]%
        {mahdisoltani2014yago3}
\bibfield{author}{\bibinfo{person}{Farzaneh Mahdisoltani},
  \bibinfo{person}{Joanna Biega}, {and} \bibinfo{person}{Fabian Suchanek}.}
  \bibinfo{year}{2014}\natexlab{}.
\newblock \showarticletitle{Yago3: A knowledge base from multilingual
  wikipedias}. In \bibinfo{booktitle}{\emph{7th biennial conference on
  innovative data systems research}}. CIDR Conference.
\newblock


\bibitem[Mao et~al\mbox{.}(2021)]%
        {mao2021negative}
\bibfield{author}{\bibinfo{person}{Xin Mao}, \bibinfo{person}{Wenting Wang},
  \bibinfo{person}{Yuanbin Wu}, {and} \bibinfo{person}{Man Lan}.}
  \bibinfo{year}{2021}\natexlab{}.
\newblock \showarticletitle{Are Negative Samples Necessary in Entity Alignment?
  An Approach with High Performance, Scalability and Robustness}. In
  \bibinfo{booktitle}{\emph{Proceedings of the 30th ACM International
  Conference on Information \& Knowledge Management}}.
  \bibinfo{pages}{1263--1273}.
\newblock


\bibitem[Mao et~al\mbox{.}(2020a)]%
        {mao2020mraea}
\bibfield{author}{\bibinfo{person}{Xin Mao}, \bibinfo{person}{Wenting Wang},
  \bibinfo{person}{Huimin Xu}, \bibinfo{person}{Man Lan}, {and}
  \bibinfo{person}{Yuanbin Wu}.} \bibinfo{year}{2020}\natexlab{a}.
\newblock \showarticletitle{MRAEA: an efficient and robust entity alignment
  approach for cross-lingual knowledge graph}. In
  \bibinfo{booktitle}{\emph{Proceedings of the 13th International Conference on
  Web Search and Data Mining}}. \bibinfo{pages}{420--428}.
\newblock


\bibitem[Mao et~al\mbox{.}(2020b)]%
        {mao2020relational}
\bibfield{author}{\bibinfo{person}{Xin Mao}, \bibinfo{person}{Wenting Wang},
  \bibinfo{person}{Huimin Xu}, \bibinfo{person}{Yuanbin Wu}, {and}
  \bibinfo{person}{Man Lan}.} \bibinfo{year}{2020}\natexlab{b}.
\newblock \showarticletitle{Relational reflection entity alignment}. In
  \bibinfo{booktitle}{\emph{Proceedings of the 29th ACM International
  Conference on Information \& Knowledge Management}}.
  \bibinfo{pages}{1095--1104}.
\newblock


\bibitem[Pei et~al\mbox{.}(2019)]%
        {pei2019semi}
\bibfield{author}{\bibinfo{person}{Shichao Pei}, \bibinfo{person}{Lu Yu},
  \bibinfo{person}{Robert Hoehndorf}, {and} \bibinfo{person}{Xiangliang
  Zhang}.} \bibinfo{year}{2019}\natexlab{}.
\newblock \showarticletitle{Semi-supervised entity alignment via knowledge
  graph embedding with awareness of degree difference}. In
  \bibinfo{booktitle}{\emph{WWW}}. \bibinfo{pages}{3130--3136}.
\newblock


\bibitem[Pei et~al\mbox{.}(2020)]%
        {pei2020rea}
\bibfield{author}{\bibinfo{person}{Shichao Pei}, \bibinfo{person}{Lu Yu},
  \bibinfo{person}{Guoxian Yu}, {and} \bibinfo{person}{Xiangliang Zhang}.}
  \bibinfo{year}{2020}\natexlab{}.
\newblock \showarticletitle{Rea: Robust cross-lingual entity alignment between
  knowledge graphs}. In \bibinfo{booktitle}{\emph{Proceedings of the 26th ACM
  SIGKDD International Conference on Knowledge Discovery \& Data Mining}}.
  \bibinfo{pages}{2175--2184}.
\newblock


\bibitem[Shi and Xiao(2019)]%
        {shi2019modeling}
\bibfield{author}{\bibinfo{person}{Xiaofei Shi} {and} \bibinfo{person}{Yanghua
  Xiao}.} \bibinfo{year}{2019}\natexlab{}.
\newblock \showarticletitle{Modeling multi-mapping relations for precise
  cross-lingual entity alignment}. In \bibinfo{booktitle}{\emph{EMNLP}}.
  \bibinfo{pages}{813--822}.
\newblock


\bibitem[Sun et~al\mbox{.}(2017b)]%
        {iptranse}
\bibfield{author}{\bibinfo{person}{M Sun}, \bibinfo{person}{H Zhu},
  \bibinfo{person}{R Xie}, {and} \bibinfo{person}{Z Liu}.}
  \bibinfo{year}{2017}\natexlab{b}.
\newblock \showarticletitle{Iterative Entity Alignment Via Joint Knowledge
  Embeddings [C]}. In \bibinfo{booktitle}{\emph{International Joint Conference
  on Artificial Intelligence. AAAI Press}}.
\newblock


\bibitem[Sun et~al\mbox{.}(2017a)]%
        {JAPE}
\bibfield{author}{\bibinfo{person}{Zequn Sun}, \bibinfo{person}{Wei Hu}, {and}
  \bibinfo{person}{Chengkai Li}.} \bibinfo{year}{2017}\natexlab{a}.
\newblock \showarticletitle{Cross-Lingual Entity Alignment via Joint
  Attribute-Preserving Embedding}. \bibinfo{pages}{628--644}.
\newblock
\showISBNx{978-3-319-68287-7}
\urldef\tempurl%
\url{https://doi.org/10.1007/978-3-319-68288-4_37}
\showDOI{\tempurl}


\bibitem[Sun et~al\mbox{.}(2018)]%
        {sun2018bootstrapping}
\bibfield{author}{\bibinfo{person}{Zequn Sun}, \bibinfo{person}{Wei Hu},
  \bibinfo{person}{Qingheng Zhang}, {and} \bibinfo{person}{Yuzhong Qu}.}
  \bibinfo{year}{2018}\natexlab{}.
\newblock \showarticletitle{Bootstrapping Entity Alignment with Knowledge Graph
  Embedding}. In \bibinfo{booktitle}{\emph{IJCAI, vol.18}}.
  \bibinfo{pages}{4396--4402}.
\newblock


\bibitem[Sun et~al\mbox{.}(2020a)]%
        {sun2020knowledge}
\bibfield{author}{\bibinfo{person}{Zequn Sun}, \bibinfo{person}{Chengming
  Wang}, \bibinfo{person}{Wei Hu}, \bibinfo{person}{Muhao Chen},
  \bibinfo{person}{Jian Dai}, \bibinfo{person}{Wei Zhang}, {and}
  \bibinfo{person}{Yuzhong Qu}.} \bibinfo{year}{2020}\natexlab{a}.
\newblock \showarticletitle{Knowledge graph alignment network with gated
  multi-hop neighborhood aggregation}. In \bibinfo{booktitle}{\emph{Proceedings
  of the AAAI Conference on Artificial Intelligence}},
  Vol.~\bibinfo{volume}{34}. \bibinfo{pages}{222--229}.
\newblock


\bibitem[Sun et~al\mbox{.}(2020b)]%
        {sun13benchmarking}
\bibfield{author}{\bibinfo{person}{Zequn Sun}, \bibinfo{person}{Qingheng
  Zhang}, \bibinfo{person}{Wei Hu}, \bibinfo{person}{Chengming Wang},
  \bibinfo{person}{Muhao Chen}, \bibinfo{person}{Farahnaz Akrami}, {and}
  \bibinfo{person}{Chengkai Li}.} \bibinfo{year}{2020}\natexlab{b}.
\newblock \showarticletitle{A Benchmarking Study of Embedding-based Entity
  Alignment for Knowledge Graphs}.
\newblock \bibinfo{journal}{\emph{Proceedings of the VLDB Endowment}}
  \bibinfo{volume}{13}, \bibinfo{number}{11} (\bibinfo{year}{2020}).
\newblock


\bibitem[Tang et~al\mbox{.}(2006)]%
        {tang2006using}
\bibfield{author}{\bibinfo{person}{Jie Tang}, \bibinfo{person}{Juanzi Li},
  \bibinfo{person}{Bangyong Liang}, \bibinfo{person}{Xiaotong Huang},
  \bibinfo{person}{Yi Li}, {and} \bibinfo{person}{Kehong Wang}.}
  \bibinfo{year}{2006}\natexlab{}.
\newblock \showarticletitle{Using Bayesian decision for ontology mapping}.
\newblock \bibinfo{journal}{\emph{Journal of web semantics}}
  \bibinfo{volume}{4}, \bibinfo{number}{4} (\bibinfo{year}{2006}),
  \bibinfo{pages}{243--262}.
\newblock


\bibitem[Tang et~al\mbox{.}(2020)]%
        {tang2019bert-int}
\bibfield{author}{\bibinfo{person}{Xiaobin Tang}, \bibinfo{person}{Jing Zhang},
  \bibinfo{person}{Bo Chen}, \bibinfo{person}{Yang Yang}, \bibinfo{person}{Hong
  Chen}, {and} \bibinfo{person}{Cuiping Li}.} \bibinfo{year}{2020}\natexlab{}.
\newblock \showarticletitle{BERT-INT: A BERT-based Interaction Model For
  Knowledge Graph Alignment.}. In \bibinfo{booktitle}{\emph{IJCAI}}.
  \bibinfo{pages}{3174--3180}.
\newblock


\bibitem[Veli{\v{c}}kovi{\'c} et~al\mbox{.}(2018)]%
        {velivckovic2017graph}
\bibfield{author}{\bibinfo{person}{Petar Veli{\v{c}}kovi{\'c}},
  \bibinfo{person}{Guillem Cucurull}, \bibinfo{person}{Arantxa Casanova},
  \bibinfo{person}{Adriana Romero}, \bibinfo{person}{Pietro Li{\`o}}, {and}
  \bibinfo{person}{Yoshua Bengio}.} \bibinfo{year}{2018}\natexlab{}.
\newblock \showarticletitle{Graph Attention Networks}. In
  \bibinfo{booktitle}{\emph{International Conference on Learning
  Representations}}.
\newblock


\bibitem[Vrande{\v{c}}i{\'c}(2014)]%
        {vrandevcic2014wikidata}
\bibfield{author}{\bibinfo{person}{Vrande{\v{c}}i{\'c}}.}
  \bibinfo{year}{2014}\natexlab{}.
\newblock \showarticletitle{Wikidata: a free collaborative knowledgebase}.
\newblock \bibinfo{journal}{\emph{Commun. ACM}} \bibinfo{volume}{57},
  \bibinfo{number}{10} (\bibinfo{year}{2014}), \bibinfo{pages}{78--85}.
\newblock


\bibitem[Wang and Liu(2021)]%
        {wang2021understanding}
\bibfield{author}{\bibinfo{person}{Feng Wang} {and} \bibinfo{person}{Huaping
  Liu}.} \bibinfo{year}{2021}\natexlab{}.
\newblock \showarticletitle{Understanding the behaviour of contrastive loss}.
  In \bibinfo{booktitle}{\emph{Proceedings of the IEEE/CVF Conference on
  Computer Vision and Pattern Recognition}}. \bibinfo{pages}{2495--2504}.
\newblock


\bibitem[Wang et~al\mbox{.}(2013)]%
        {wang2013boosting}
\bibfield{author}{\bibinfo{person}{Zhichun Wang}, \bibinfo{person}{Juanzi Li},
  {and} \bibinfo{person}{Jie Tang}.} \bibinfo{year}{2013}\natexlab{}.
\newblock \showarticletitle{Boosting cross-lingual knowledge linking via
  concept annotation}. In \bibinfo{booktitle}{\emph{Twenty-Third International
  Joint Conference on Artificial Intelligence}}.
\newblock


\bibitem[Wang et~al\mbox{.}(2018a)]%
        {GCN-Align}
\bibfield{author}{\bibinfo{person}{Zhichun Wang}, \bibinfo{person}{Qingsong
  Lv}, \bibinfo{person}{Xiaohan Lan}, {and} \bibinfo{person}{Yu Zhang}.}
  \bibinfo{year}{2018}\natexlab{a}.
\newblock \showarticletitle{Cross-lingual Knowledge Graph Alignment via Graph
  Convolutional Networks}. \bibinfo{pages}{349--357}.
\newblock
\urldef\tempurl%
\url{https://doi.org/10.18653/v1/D18-1032}
\showDOI{\tempurl}


\bibitem[Wang et~al\mbox{.}(2018b)]%
        {wang2018cross}
\bibfield{author}{\bibinfo{person}{Zhichun Wang}, \bibinfo{person}{Qingsong
  Lv}, \bibinfo{person}{Xiaohan Lan}, {and} \bibinfo{person}{Yu Zhang}.}
  \bibinfo{year}{2018}\natexlab{b}.
\newblock \showarticletitle{Cross-lingual knowledge graph alignment via graph
  convolutional networks}. In \bibinfo{booktitle}{\emph{EMNLP}}.
  \bibinfo{pages}{349--357}.
\newblock


\bibitem[Wang et~al\mbox{.}(2020)]%
        {wang2020knowledge}
\bibfield{author}{\bibinfo{person}{Zhichun Wang}, \bibinfo{person}{Jinjian
  Yang}, {and} \bibinfo{person}{Xiaoju Ye}.} \bibinfo{year}{2020}\natexlab{}.
\newblock \showarticletitle{Knowledge graph alignment with entity-pair
  embedding}. In \bibinfo{booktitle}{\emph{Proceedings of the 2020 Conference
  on Empirical Methods in Natural Language Processing (EMNLP)}}.
  \bibinfo{pages}{1672--1680}.
\newblock


\bibitem[Wu et~al\mbox{.}(2019b)]%
        {wu2019relation}
\bibfield{author}{\bibinfo{person}{Yuting Wu}, \bibinfo{person}{Xiao Liu},
  \bibinfo{person}{Yansong Feng}, \bibinfo{person}{Zheng Wang},
  \bibinfo{person}{Rui Yan}, {and} \bibinfo{person}{Dongyan Zhao}.}
  \bibinfo{year}{2019}\natexlab{b}.
\newblock \showarticletitle{Relation-Aware Entity Alignment for Heterogeneous
  Knowledge Graphs}. IJCAI.
\newblock


\bibitem[Wu et~al\mbox{.}(2019a)]%
        {wu2019jointly}
\bibfield{author}{\bibinfo{person}{Yuting Wu}, \bibinfo{person}{Xiao Liu},
  \bibinfo{person}{Yansong Feng}, \bibinfo{person}{Zheng Wang}, {and}
  \bibinfo{person}{Dongyan Zhao}.} \bibinfo{year}{2019}\natexlab{a}.
\newblock \showarticletitle{Jointly Learning Entity and Relation
  Representations for Entity Alignment}. In
  \bibinfo{booktitle}{\emph{Proceedings of the 2019 Conference on Empirical
  Methods in Natural Language Processing and the 9th International Joint
  Conference on Natural Language Processing (EMNLP-IJCNLP)}}.
  \bibinfo{pages}{240--249}.
\newblock


\bibitem[Xu et~al\mbox{.}(2019)]%
        {xu2019cross-lingual}
\bibfield{author}{\bibinfo{person}{Kun Xu}, \bibinfo{person}{Liwei Wang},
  \bibinfo{person}{Mo Yu}, \bibinfo{person}{Yansong Feng}, \bibinfo{person}{Yan
  Song}, \bibinfo{person}{Zhiguo Wang}, {and} \bibinfo{person}{Dong Yu}.}
  \bibinfo{year}{2019}\natexlab{}.
\newblock \showarticletitle{Cross-lingual Knowledge Graph Alignment via Graph
  Matching Neural Network}. In \bibinfo{booktitle}{\emph{Proceedings of the
  57th Annual Meeting of the Association for Computational Linguistics}}.
  \bibinfo{pages}{3156--3161}.
\newblock


\bibitem[Yang et~al\mbox{.}(2019)]%
        {yang2019aligning}
\bibfield{author}{\bibinfo{person}{Hsiu-Wei Yang}, \bibinfo{person}{Yanyan
  Zou}, \bibinfo{person}{Peng Shi}, \bibinfo{person}{Wei Lu},
  \bibinfo{person}{Jimmy Lin}, {and} \bibinfo{person}{Xu Sun}.}
  \bibinfo{year}{2019}\natexlab{}.
\newblock \showarticletitle{Aligning Cross-Lingual Entities with Multi-Aspect
  Information}. In \bibinfo{booktitle}{\emph{Proceedings of the 2019 Conference
  on Empirical Methods in Natural Language Processing and the 9th International
  Joint Conference on Natural Language Processing (EMNLP-IJCNLP)}}.
  \bibinfo{pages}{4431--4441}.
\newblock


\bibitem[Yu et~al\mbox{.}(2021)]%
        {yu2021generalized}
\bibfield{author}{\bibinfo{person}{Donghan Yu}, \bibinfo{person}{Yiming Yang},
  \bibinfo{person}{Ruohong Zhang}, {and} \bibinfo{person}{Yuexin Wu}.}
  \bibinfo{year}{2021}\natexlab{}.
\newblock \showarticletitle{Generalized multi-relational graph convolution
  network}. In \bibinfo{booktitle}{\emph{The World Wide Web Conference}}.
\newblock


\bibitem[Zeng et~al\mbox{.}(2021a)]%
        {zeng2021comprehensive}
\bibfield{author}{\bibinfo{person}{Kaisheng Zeng}, \bibinfo{person}{Chengjiang
  Li}, \bibinfo{person}{Lei Hou}, \bibinfo{person}{Juanzi Li}, {and}
  \bibinfo{person}{Ling Feng}.} \bibinfo{year}{2021}\natexlab{a}.
\newblock \showarticletitle{A comprehensive survey of entity alignment for
  knowledge graphs}.
\newblock \bibinfo{journal}{\emph{AI Open}}  \bibinfo{volume}{2}
  (\bibinfo{year}{2021}), \bibinfo{pages}{1--13}.
\newblock


\bibitem[Zeng et~al\mbox{.}(2021b)]%
        {zeng2021encoding}
\bibfield{author}{\bibinfo{person}{Kaisheng Zeng}, \bibinfo{person}{Chengjiang
  Li}, \bibinfo{person}{Yan Qi}, \bibinfo{person}{Xin Lv}, \bibinfo{person}{Lei
  Hou}, \bibinfo{person}{Guozheng Peng}, \bibinfo{person}{Juanzi Li}, {and}
  \bibinfo{person}{Ling Feng}.} \bibinfo{year}{2021}\natexlab{b}.
\newblock \showarticletitle{Encoding the Meaning Triangle (Object, Entity, and
  Concept) as the Semantic Foundation for Entity Alignment}. In
  \bibinfo{booktitle}{\emph{International Conference on Web Information Systems
  Engineering}}. Springer, \bibinfo{pages}{227--241}.
\newblock


\bibitem[Zeng et~al\mbox{.}(2021c)]%
        {zeng2021reinforced}
\bibfield{author}{\bibinfo{person}{Weixin Zeng}, \bibinfo{person}{Xiang Zhao},
  \bibinfo{person}{Jiuyang Tang}, {and} \bibinfo{person}{Changjun Fan}.}
  \bibinfo{year}{2021}\natexlab{c}.
\newblock \showarticletitle{Reinforced Active Entity Alignment}. In
  \bibinfo{booktitle}{\emph{Proceedings of the 30th ACM International
  Conference on Information \& Knowledge Management}}.
  \bibinfo{pages}{2477--2486}.
\newblock


\bibitem[Zeng et~al\mbox{.}(2019)]%
        {CEAFF}
\bibfield{author}{\bibinfo{person}{Weixin Zeng}, \bibinfo{person}{Xiang Zhao},
  \bibinfo{person}{Jiuyang Tang}, {and} \bibinfo{person}{Xuemin Lin}.}
  \bibinfo{year}{2019}\natexlab{}.
\newblock \showarticletitle{Collective Embedding-based Entity Alignment via
  Adaptive Features}.
\newblock


\bibitem[Zeng et~al\mbox{.}(2020)]%
        {zeng2020degree}
\bibfield{author}{\bibinfo{person}{Weixin Zeng}, \bibinfo{person}{Xiang Zhao},
  \bibinfo{person}{Wei Wang}, \bibinfo{person}{Jiuyang Tang}, {and}
  \bibinfo{person}{Zhen Tan}.} \bibinfo{year}{2020}\natexlab{}.
\newblock \showarticletitle{Degree-aware alignment for entities in tail}. In
  \bibinfo{booktitle}{\emph{Proceedings of the 43rd International ACM SIGIR
  Conference on Research and Development in Information Retrieval}}.
  \bibinfo{pages}{811--820}.
\newblock


\bibitem[Zhang et~al\mbox{.}(2019)]%
        {zhang2019multi}
\bibfield{author}{\bibinfo{person}{Qingheng Zhang}, \bibinfo{person}{Zequn
  Sun}, \bibinfo{person}{Wei Hu}, \bibinfo{person}{Muhao Chen},
  \bibinfo{person}{Lingbing Guo}, {and} \bibinfo{person}{Yuzhong Qu}.}
  \bibinfo{year}{2019}\natexlab{}.
\newblock \showarticletitle{Multi-view Knowledge Graph Embedding for Entity
  Alignment}. In \bibinfo{booktitle}{\emph{IJCAI}}.
\newblock


\bibitem[Zhu et~al\mbox{.}(2017)]%
        {zhu2017iterative}
\bibfield{author}{\bibinfo{person}{Hao Zhu}, \bibinfo{person}{Ruobing Xie},
  \bibinfo{person}{Zhiyuan Liu}, {and} \bibinfo{person}{Maosong Sun}.}
  \bibinfo{year}{2017}\natexlab{}.
\newblock \showarticletitle{Iterative Entity Alignment via Joint Knowledge
  Embeddings.}. In \bibinfo{booktitle}{\emph{IJCAI}},
  Vol.~\bibinfo{volume}{17}. \bibinfo{pages}{4258--4264}.
\newblock


\bibitem[Zhu et~al\mbox{.}(2020)]%
        {zhu2020collective}
\bibfield{author}{\bibinfo{person}{Qi Zhu}, \bibinfo{person}{Hao Wei},
  \bibinfo{person}{Bunyamin Sisman}, \bibinfo{person}{Da Zheng},
  \bibinfo{person}{Christos Faloutsos}, \bibinfo{person}{Xin~Luna Dong}, {and}
  \bibinfo{person}{Jiawei Han}.} \bibinfo{year}{2020}\natexlab{}.
\newblock \showarticletitle{Collective multi-type entity alignment between
  knowledge graphs}. In \bibinfo{booktitle}{\emph{Proceedings of The Web
  Conference 2020}}. \bibinfo{pages}{2241--2252}.
\newblock


\bibitem[Zhu et~al\mbox{.}(2019)]%
        {zhu2019neighborhood}
\bibfield{author}{\bibinfo{person}{Qiannan Zhu}, \bibinfo{person}{Xiaofei
  Zhou}, \bibinfo{person}{Jia Wu}, \bibinfo{person}{Jianlong Tan}, {and}
  \bibinfo{person}{Li Guo}.} \bibinfo{year}{2019}\natexlab{}.
\newblock \showarticletitle{Neighborhood-Aware Attentional Representation for
  Multilingual Knowledge Graphs.}. In \bibinfo{booktitle}{\emph{IJCAI}}.
  \bibinfo{pages}{1943--1949}.
\newblock


\bibitem[Zhu et~al\mbox{.}(2021)]%
        {zhu2020relation}
\bibfield{author}{\bibinfo{person}{Yao Zhu}, \bibinfo{person}{Hongzhi Liu},
  \bibinfo{person}{Zhonghai Wu}, {and} \bibinfo{person}{Yingpeng Du}.}
  \bibinfo{year}{2021}\natexlab{}.
\newblock \showarticletitle{Relation-Aware Neighborhood Matching Model for
  Entity Alignment}. In \bibinfo{booktitle}{\emph{Proceedings of the AAAI
  Conference on Artificial Intelligence}}, Vol.~\bibinfo{volume}{35}.
  \bibinfo{pages}{4749--4756}.
\newblock


\end{thebibliography}
